\documentclass[lettersize,journal]{IEEEtran}
\usepackage{times}

\usepackage[numbers]{natbib}
\usepackage{amsmath,amssymb,amsfonts}
\usepackage[ruled,linesnumbered, noend]{algorithm2e}
\usepackage{multicol}
\usepackage[bookmarks=true]{hyperref}
\usepackage{booktabs}
\usepackage{multirow}
\usepackage{threeparttable}
\usepackage{siunitx}
\usepackage{array}
\usepackage[table]{xcolor}
\usepackage{graphicx}
\usepackage{gensymb}
\usepackage{optidef}
\usepackage{cuted} 
\usepackage{capt-of}
\newcolumntype{R}{S[table-format=3.2]} 
\newcolumntype{E}{S[table-format=1.3]} 
\SetKwComment{Comment}{$\triangleright$\ }{}

\begin{document}

\title{\textit{NavRL++}: A System-Level Framework for Improving Sim-to-Real Transfer in Reinforcement Learning-Based Robot Navigation}

\author{Zhefan Xu, Hanyu Jin, and Kenji Shimada
\thanks{Zhefan Xu, Hanyu Jin, and Kenji Shimada are with the Department of Mechanical Engineering, Carnegie Mellon University, 5000 Forbes Ave, Pittsburgh, PA, 15213, USA.
        {\tt\small zhefanx@andrew.cmu.edu}}%
}

\maketitle
\begin{strip}
    \centering
    \vspace{-1.0cm}
    \includegraphics[width=1.0\textwidth]{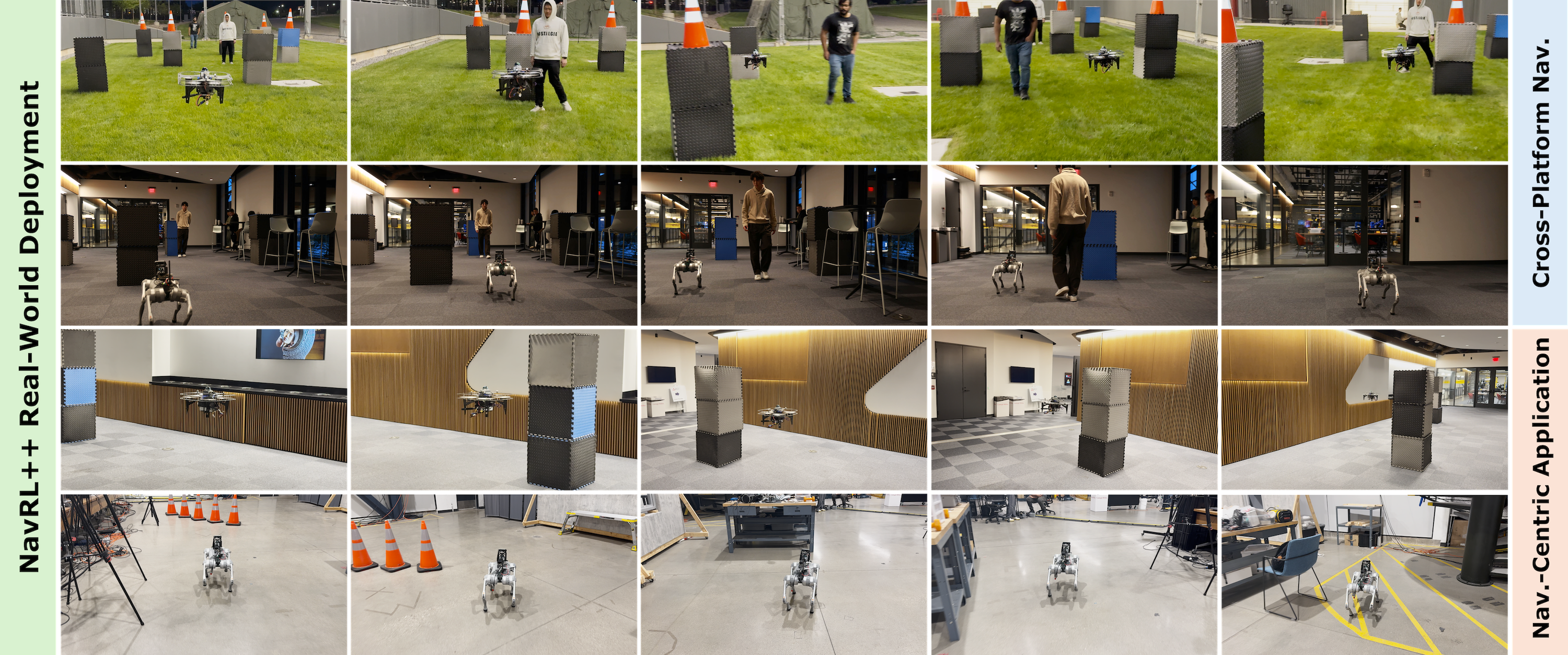}
    \captionof{figure}{Overview of \textit{NavRL++} real-world deployment. The proposed system supports zero-shot deployment across multiple environments and robotic platforms, including quadruped and aerial robots. The results highlight robust navigation in dynamic, cluttered, and task-oriented scenarios such as pedestrian collision avoidance, surface inspection, and unknown exploration, demonstrating deployment across multiple robotic platforms and real-world navigation scenarios.}
    \label{fig:intro_figure}
\end{strip}
\begin{abstract}
Recent years have witnessed significant progress in autonomous navigation using reinforcement learning. However, existing approaches largely emphasize reinforcement learning framework design, such as input representations, action spaces, and reward functions, while providing limited analysis of sim-to-real transfer and insufficient insight into how training strategies affect real-world deployment performance. To bridge this gap, we not only introduce an effective RL framework but also present a complete training and deployment pipeline, along with a systematic empirical study that disentangles the key factors affecting sim-to-real transfer in reinforcement learning-based navigation, including sensor noise, perception failures, system latency, and control response. Building on insights from this analysis, we introduce perturbation-aware fine-tuning, a post-training adaptation strategy that improves transfer robustness by explicitly accounting for empirically identified domain discrepancies. To further mitigate perception degradation and enhance control smoothness in real-world deployment, we propose a Transformer-based temporal reasoning policy that leverages short-horizon observation for navigation control. We quantitatively evaluate how individual sim-to-real perturbations and training design choices impact navigation performance across environments. Experimental results demonstrate that the proposed training strategy and policy architecture outperform learning-based baselines in both static and dynamic environments, while achieving performance comparable to optimization-based planners in static settings. We validate our approach through real-world deployment on multiple robotic platforms, including aerial and legged robots, across navigation-centric tasks such as exploration and inspection, demonstrating zero-shot sim-to-real transfer. \end{abstract}

\IEEEpeerreviewmaketitle

\section{Introduction}
Safe navigation is a fundamental capability of autonomous mobile robots, enabling tasks such as inspection \cite{inspection_example}, surveillance \cite{survalience_example}, and exploration \cite{racer}. Prior works \cite{fast-planner}\cite{ego-planner}\cite{far_planner}\cite{fapp} have demonstrated safe and efficient navigation in specific environments using optimization-based and handcrafted algorithms. More recently, deep reinforcement learning has emerged as a promising alternative due to its ability to learn navigation behaviors directly from interaction across diverse environments. With the rapid development of reinforcement learning frameworks, identifying effective training recipes for reinforcement learning–based navigation has become increasingly critical.

Reinforcement learning–based navigation has been extensively studied in recent years, but several fundamental challenges still remain unresolved. Major challenges lie in the intrinsic difficulties of reinforcement learning itself, such as sample inefficiency, sparse rewards, and high gradient variance \cite{sutton1999policy}\cite{duan2016benchmarking}, which often make navigation policies difficult to converge. To mitigate these issues in robot navigation, prior work has explored strategies such as introducing a human-guided learning stage \cite{hu2025toward}, carefully shaping reward functions \cite{he2024agile}, and leveraging highly parallelized simulation environments \cite{isaaclab} to improve training efficiency and convergence. Besides, when deploying these methods to real-world systems, a second major challenge emerges: sim-to-real transfer, where a large discrepancy exists between simulation and the real world. To address this gap, many approaches \cite{loquercio2021learning}\cite{drl-vo}\cite{yopo}\cite{NavRL} carefully design the state and action representations. One common strategy \cite{drl-vo}\cite{NavRL} is to use state based inputs instead of raw sensor data and other approaches \cite{loquercio2021learning}\cite{yopo} predicts pre-generated dynamics or trajectories as outputs. These designs aim to reduce the discrepancy between simulation and reality. While these methods demonstrate promising results, there is still a lack of systematic and quantitative analysis of how different factors contribute to sim-to-real discrepancies and which training recipes lead to optimal navigation performance. We hypothesize that understanding the relative impact of these deployment discrepancies is critical for improving the reliability of learned navigation policies in real-world environments.

To address these limitations, we propose a reinforcement learning-based navigation framework that includes training and deployment. Built upon this framework, we conduct a comprehensive empirical study that disentangles the key factors affecting sim-to-real transfer in reinforcement learning–based navigation. Our study examines how mismatches introduced at different stages, ranging from sensing and perception to action generation and control execution, affect navigation performance. Leveraging the insights from this analysis, we propose perturbation-aware fine-tuning, a post-training adaptation approach that enhances transfer robustness by explicitly incorporating observed domain mismatches. In addition, to reduce perception-induced degradation and achieve smoother control behavior, we introduce a Transformer-based temporal reasoning architecture that utilizes short-horizon observation histories for navigation control. Finally, we evaluate the deployment robustness of the learned policies on real-world navigation-centric tasks, including robotic inspection and autonomous exploration across multiple robotic platforms. The contributions are summarized as follows:

\begin{itemize}
    \item \textbf{Comprehensive RL Navigation Framework.} We introduce \textit{NavRL++}, a complete reinforcement learning-based navigation framework that spans simulation training and real-world deployment, covering the full pipeline from perception to control. The framework will be released on GitHub to support reproducibility and future research.
    \item \textbf{Perturbation-Aware Sim-to-Real Adaptation.} We introduce a perturbation-aware fine-tuning paradigm that adapts pretrained reinforcement learning policies using empirically identified domain mismatches arising from sensing, perception, latency, and control discrepancies.
    \item \textbf{Temporal Reasoning Under Deployment Uncertainty.} We show that short-horizon temporal reasoning improves robustness to delayed and imperfect observations while reducing control oscillation during deployment.
    \item \textbf{Multi-modal Sensor Support.} We design an extensible and lightweight perception module within our navigation framework that supports multi-modal inputs, including RGB-D cameras, LiDAR, and their combinations.
    \item \textbf{Factorized Empirical Analysis of Sim-to-Real Transfer.} We present an empirical analysis that systematically disentangles key sim-to-real perturbations and training design choices, and quantifies their effects on navigation performance across environments with varying difficulty.
    \item \textbf{Cross-Task Real-World Deployment.} We validate the proposed approach through real-world deployment on multiple navigation-centric tasks across different robotic platforms, demonstrating zero-shot sim-to-real transfer. 
\end{itemize}

\begin{figure}[t] 
    \centering
    \includegraphics[scale=1.13]{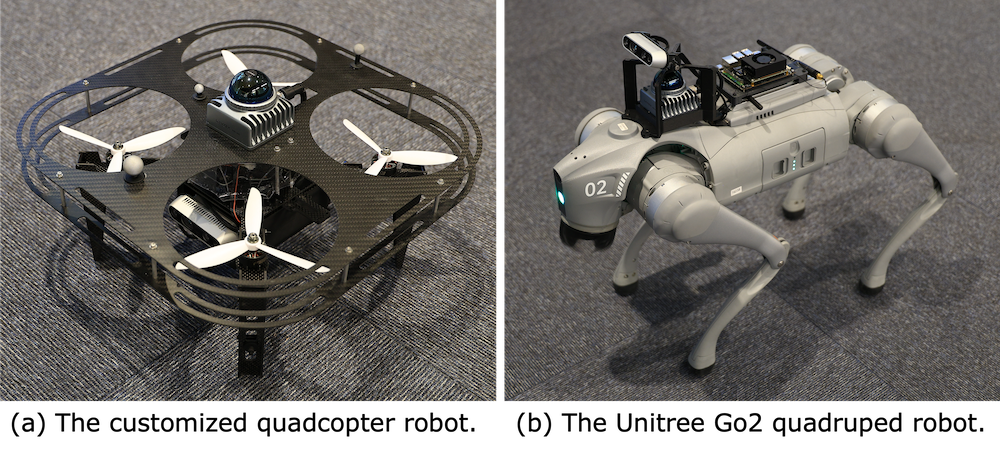}
    \caption{Illustration of the robotic platforms used to deploy our RL navigation framework. Each platform is equipped with a lightweight onboard LiDAR (Livox Mid-360) and an RGB-D camera (Intel RealSense D435i). }
    \label{fig:robot_platform}
\end{figure}

\section{Related Work}
Autonomous robot navigation has been extensively studied using handcrafted and optimization-based methods \cite{fast-planner}\cite{ego-planner}\cite{far_planner}\cite{fapp}, which have demonstrated promising performance in many scenarios. Recently, learning-based navigation methods have shown strong potential to further improve adaptability and generalization across diverse environments. However, they still suffer from several fundamental challenges, including data scarcity and sim-to-real transfer. This section mainly discusses learning-based navigation approaches, including two major categories: supervised learning methods and reinforcement learning methods. We then review sim-to-real transfer techniques that have been used in navigation applications.

\textbf{Supervised Learning Methods:} Methods in this category rely on real-world or synthetic data to train navigation policies. Early approaches \cite{kim2015deep}\cite{16supervised}\cite{smolyanskiy2017toward} use manually labeled real-world image data to learn high-level navigation commands, while others \cite{gandhi2017learning}\cite{padhy2018deep}\cite{khan2018learning} predict collision likelihood to guide navigation. To address data scarcity, simulation-based training is explored in \cite{dionisio2018deep}, alleviating data collection costs but introducing sim-to-real transfer challenges. Loquercio et al. \cite{loquercio2018dronet} augment manually collected datasets with large-scale autonomous driving data, improving outdoor deployment performance, and further extend this idea by learning waypoint prediction for fast trajectory generation \cite{loquercio2021learning}. Tolani et al. \cite{tolani2021visual} reduce labeling effort by using a model predictive controller as an expert for automatic data annotation. More recently, navigation foundation models \cite{shah2023vint}\cite{sridhar2024nomad} leverage large-scale real-world robotic datasets to train diffusion-based policies for safe trajectory prediction, achieving strong generalization. Gradient-based training methods are introduced in \cite{yopo}\cite{zhang2025learning}, while Han et al. \cite{han2025neupan} propose a model-based learning approach for end-to-end trajectory generation from LiDAR inputs. In \cite{peng2025logoplanner}, an implicit state estimation module is incorporated to reduce reliance on explicit localization. Several works \cite{jung2018perception}\cite{simon2023mononav}\cite{kulkarni2023task} adopt indirect navigation strategies by learning perception modules that predict collisions, objects, or distances to guide navigation decisions. Despite promising performance, supervised learning-based navigation methods typically require large-scale data collection or carefully curated synthetic datasets. In contrast, our work focuses on reinforcement learning, which enables robots to directly learn navigation behaviors through interaction without extensive manual data curation.

\textbf{Reinforcement Learning Methods:} Unlike supervised learning, reinforcement learning (RL)-based methods learn navigation behaviors through interaction with the environment. Early works \cite{sadeghi2016cad2rl}\cite{xie2017towards}\cite{chen2017decentralized}\cite{chen2017socially}\cite{roghair2021vision}\cite{singla2019memory} primarily adopt value-based approaches that map sensor observations to discrete actions, demonstrating effective navigation but suffering from suboptimality due to limited action resolution. More recent studies increasingly favor policy-based methods using the actor–critic paradigm, which enable more stable training and support continuous action spaces. For example, Gao et al. \cite{gao2021vision} fuse semantic and depth information from RGB images to train an actor–critic policy for indoor navigation. In \cite{kaufmann2023champion}, a PPO-based policy \cite{ppo} achieves superhuman performance in drone racing. Xie et al. \cite{xie2023drl} incorporate velocity-obstacle-based reward design for wheeled robot navigation, while related works \cite{wang2025end}\cite{qi2025distributed} employ control or position barrier functions to encourage collision avoidance. End-to-end RL policies that map LiDAR inputs directly to action commands are explored in \cite{wang2025omni}\cite{xu2025flow} for quadruped and aerial robots. He et al. \cite{he2024agile} further demonstrate agile quadruped collision avoidance by combining learned policies with failure recovery mechanisms, and Xu et al. \cite{NavRL} adopt a safety shield to enable navigation in real-world dynamic environments. Similar policy-based approaches have been investigated with diverse state, action, and reward formulations \cite{song2021vision}\cite{xue2021vision}\cite{kim2022towards}\cite{brilli2023monocular}. Additionally, some works \cite{krinner2024time}\cite{romero2024actor} use reinforcement learning to optimize control parameters rather than directly learning navigation policies. Despite extensive exploration of RL training design, ranging from state, action, and reward representations, few works systematically examine how sim-to-real discrepancies impact deployment performance or identify training recipes that enable strong zero-shot sim-to-real transfer. In addition to introducing a new RL framework, this work empirically analyzes key factors affecting sim-to-real transfer and identifies effective training parameters for improved real-world performance.

\textbf{Sim-to-Real Transfer Techniques:} Simulation-to-reality transfer remains a fundamental and unresolved challenge in reinforcement learning, affecting a wide range of robotic tasks including grasping \cite{liu2022digital}\cite{jiang2024transic}, manipulation \cite{scheikl2022sim}, and navigation \cite{wu2023human}. A primary source of sim-to-real discrepancy arises from mismatches in sensor observations. To address this issue, several works \cite{xie2017towards}\cite{gao2021vision} use depth prediction modules to transform RGB images into a consistent depth representation, thereby reducing perceptual gaps. Other approaches adopt state-based inputs \cite{everett2021collision}\cite{NavRL} to avoid large discrepancies in high-dimensional image spaces. Song et al. \cite{song2022learning} further distill state-based policies trained in simulation into image-based policies for real-world deployment, while \cite{hoeller2021learning} trains an image encoder to generate latent representations suitable for policy learning. Contrastive learning has also been explored to improve visual robustness and scene transfer \cite{xing2024contrastive}. Beyond perception, digital twin construction \cite{liu2022digital} has been proposed to improve simulation fidelity for grasping tasks, while human-in-the-loop methods \cite{wu2023human}\cite{jiang2024transic}\cite{hu2025toward} enable policy correction during training and deployment. Domain randomization \cite{tiboni2023dropo} further aims to expose policies to a broad distribution of conditions to improve robustness in real-world environments.In contrast to prior approaches that primarily target isolated perception or policy-level discrepancies, this work studies deployment robustness from a system-level autonomy perspective spanning sensing, perception, latency, and control execution. We systematically investigate sim-to-real gaps across the full navigation stack, spanning sensing, perception, and control, and introduce a post-training adaptation method that enable learned policies to adapt to a wide range of real-world conditions. This perspective provides a complementary and broadly applicable solution that can be integrated with existing sim-to-real techniques. In addition, we present a reinforcement learning training framework with exploration of key training parameters, offering practical guidance for minimizing sim-to-real gaps and achieving reliable real-world navigation.

\section{Problem Formulation}
Navigation for mobile robots refers to the task of moving from one location to another using onboard sensing and computation. While in some contexts it denotes long-horizon planning with access to a global map, in this work it specifically refers to local navigation in cluttered environments,  emphasizing both collision avoidance and goal-directed motion.

\textbf{Robot Platforms and Sensors.} This work targets general mobile robotic platforms and conducts experiments on both quadcopter UAVs and quadruped legged robots (Fig. \ref{fig:robot_platform}). These platforms are typically equipped with RGB-D cameras, LiDAR sensors, or a combination of both to perceive the environments. Our objective is to develop a unified policy network that enables autonomous navigation using onboard sensing and computation across diverse robotic embodiments.

\textbf{Formulation.} We model the navigation using a Markov Decision Process (MDP) framework, represented by the tuple $(S, A, P, R, \gamma)$. Here, $S$ denotes the state space, including the robot’s internal states and sensory observations, and $A$ represents the action space. The system dynamics are captured by the transition model $P(s_{t+1} \mid s_t, a_t)$, that describes how the environment evolves given the robot’s actions. The reward function $R(s_t, a_t)$ is designed to promote goal-directed motion while discouraging collisions and inefficient behaviors. The objective is to learn an optimal policy $\pi^{*}(a_t \mid s_t)$ that maximizes the expected cumulative discounted reward:
\begin{equation}
    \pi^* = \arg \max_{\pi} \mathbb{E} \left[ \sum_{t=0}^{T} \gamma^t R(s_t, a_t) \right],
\end{equation}
where $\gamma \in [0, 1]$ is the discount factor for future rewards. This formulation allows reinforcement learning to be applied for training robots to navigate safely in dynamic environments.

\begin{figure*}[t]
    \centering
    \includegraphics[scale=0.92]{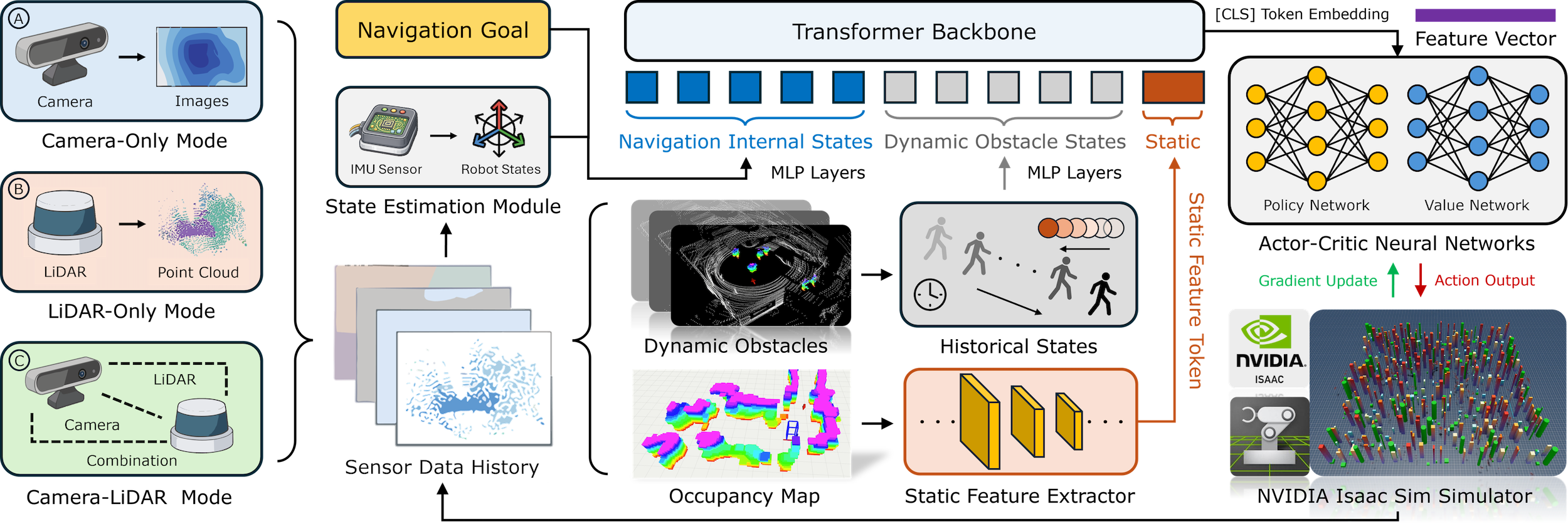}
    \caption{Overview of the proposed reinforcement learning–based autonomous navigation framework. The system supports camera-only, LiDAR-only, and camera–LiDAR fusion modes. Sensor data are processed by a state estimation module and aggregated as sensor history to form navigation internal states, while a perception module (not shown) extracts static occupancy maps and dynamic obstacle information from the environment. A transformer backbone encodes these multi-modal inputs, and the extracted feature vector is fed into actor–critic neural networks to generate control actions and update the policy via proximal policy optimization. The framework is trained and validated in the NVIDIA Isaac Sim environment before real-world deployment. }
    \label{fig:framework overview}
\end{figure*}

\section{Methodology}
This section presents the methodology for training and deploying the navigation policy. Sec. \ref{sec: framework} introduces the overall reinforcement learning framework for navigation. Sec. \ref{sec: RL formulation} details the reinforcement learning formulation, including the definitions of states, actions, and rewards. Sec. \ref{sec:network_design} presents the temporal reasoning network architecture, while Sec. \ref{sec: training strategy} describes the training strategy. Finally, Sec. \ref{sec:deployment} outlines the real-world deployment strategy.

\subsection{RL Training Framework} \label{sec: framework}
The proposed navigation system is illustrated in Fig. \ref{fig:framework overview}. In contrast to prior works \cite{yopo}\cite{NavRL}\cite{zhang2025learning}, the proposed system supports three input modalities: (1) camera-only mode providing RGB-D images, (2) LiDAR-only mode providing 3D point cloud measurements, and (3) a camera–LiDAR fusion mode that combines both visual and geometric information. These sensing configurations reflect commonly deployed sensing modalities in mobile robotic systems.

To reduce modality-specific deployment discrepancies across sensing configurations, we incorporate a state estimation module and a perception module (the latter not shown in the figure and detailed in Sec. \ref{sec:deployment}). The state estimation module, built upon either vision-based \cite{vins} or LiDAR-based \cite{fast-lio2} methods, estimates the robot’s pose and velocity in real time. These quantities are combined with the user-defined navigation goal to form the navigation internal state.

The perception module processes sensor data history to extract structured representations of the environment, including static occupancy information and dynamic obstacle states. These static and dynamic features are encoded separately and provided as inputs to the transformer backbone. The transformer adopts an attention mechanism to integrate multimodal inputs and generate a compact feature representation.

Policy learning is performed using the Proximal Policy Optimization (PPO) algorithm \cite{ppo}. The feature corresponding to the \texttt{[CLS]} token from the transformer backbone is fed into multi-layer perceptron (MLP)-based policy and value networks under an actor–critic framework. Training is conducted entirely in simulation, where sensor data are directly accessible from the simulator. During real-world deployment, the perception module replaces simulated observations and processes multimodal sensor inputs to ensure policy execution.

\subsection{Reinforcement Learning Formulation} \label{sec: RL formulation}
This section presents the detailed reinforcement learning formulation, including state representation, action parameterization, and reward design. We emphasize that a well-structured formulation not only improves training stability and performance but also enhances sim-to-real transferability. Based on our empirical observations, inappropriate combinations of state, action, and reward definitions can significantly degrade learning performance or even prevent convergence altogether.

\textbf{Input State Representation.} The input state must encode both environmental context and the robot’s internal information to effectively guide navigation. Using raw sensor data is the most direct approach, as it preserves full environmental information. However, for camera-based inputs, discrepancies between simulation and real-world visual domains introduce significant sim-to-real transfer challenges. Although advanced preprocessing methods (e.g., \cite{hoeller2021learning}) can mitigate this issue, they typically require substantial training effort and costly real-world data collection. For LiDAR inputs, raw point clouds must be processed and downsampled to satisfy real-time computational constraints. Furthermore, different sensor configurations often necessitate separate model training pipelines, reducing efficiency and scalability. To address these limitations, we adopt a distance-based representation for static obstacles and structured state representations for dynamic obstacles. This unified formulation reduces sensor modality-specific discrepancies and promotes more consistent behavior between simulation and real-world deployment.

\begin{figure}[t] 
    \centering
    \includegraphics[scale=0.51]{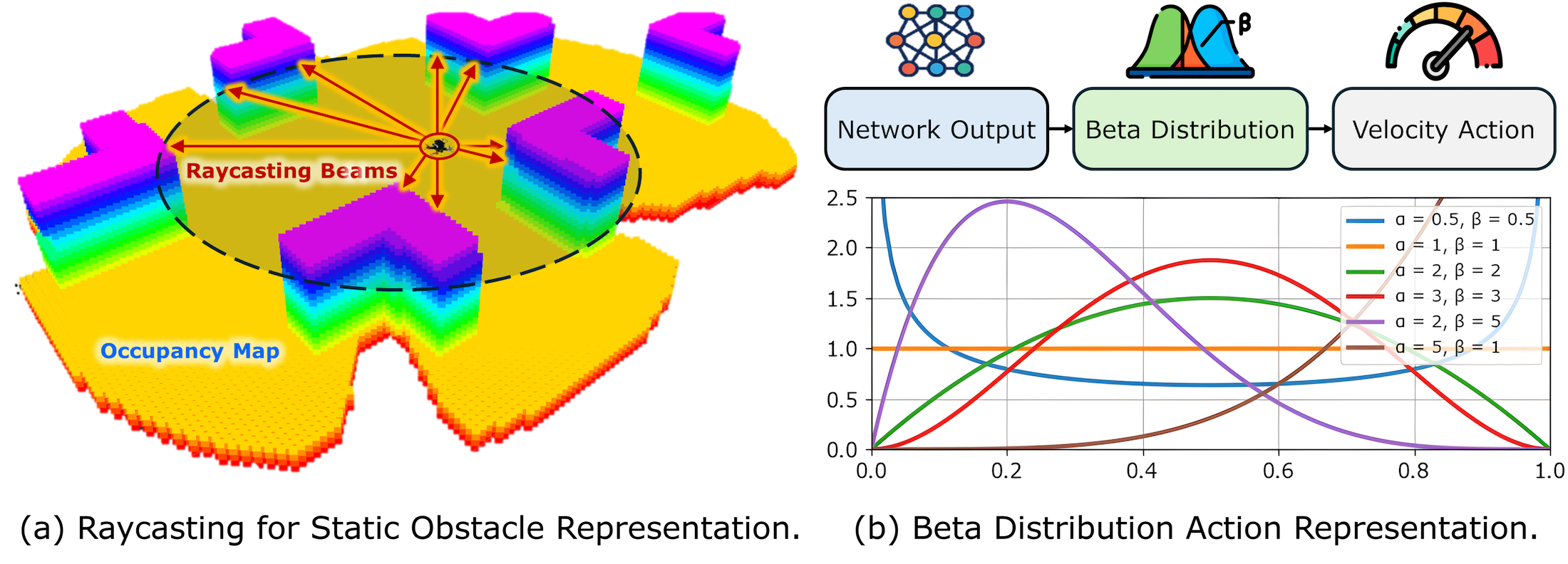}
    \caption{Illustration of static obstacle and action representation. (a) Occupancy map with raycasting beams encodes surrounding geometry. (b) Network outputs Beta distribution parameters to generate continuous velocity actions.}
    \label{fig:formulation_representation}
\end{figure}

In our state design, we explicitly distinguish between static and dynamic obstacles due to their fundamentally different properties. Static obstacles are stationary and may exhibit arbitrarily complex geometries, whereas dynamic obstacles are typically modeled as rigid bodies characterized by their positions and velocities. To represent static structures, we adopt a voxel-based local map that captures complex geometries while efficiently maintaining historical spatial information. However, directly using dense occupancy grids as neural network input is computationally inefficient. Therefore, instead of feeding occupancy values directly into the policy, we perform ego-centric ray casting from the robot's position into the local voxel map, as illustrated in Fig. \ref{fig:formulation_representation}a. The distances measured along these rays are arranged in a structured array to form a compact representation of surrounding geometry. Specifically, the horizontal plane is discretized at a resolution of 10 degree (36 beams), and the vertical plane is discretized from -10 to 20 degree at 10\degree intervals (4 beams), resulting in 144 distance measurements. These parameters are selected to balance representation capability and computational efficiency. The static obstacle state is represented as:
\begin{equation} \label{eqn: static states}
    \mathcal{S}_{\text{static}} = [\mathcal{S}_{h, 1}, \ldots,\mathcal{S}_{h, 36}], \ \  \mathcal{S}_{h, i} = [\mathcal{D}_{h, i}^{v, 1}, \ldots, \mathcal{D}_{h, i}^{v, 4}],
\end{equation}
where $\mathcal{D}_{h, i}^{v, j}$ denotes the raycasting distance to the nearest obstacle along the corresponding beam. In implementation, rather than using absolute distances, we define a maximum local sensing range (4 m in our experiments) and encode distances relative to this limit. Rays that exceed the maximum range are truncated, producing zero-valued entries after normalization, thereby preventing distant obstacles from disproportionately influencing the policy. We note that the beam resolution and sensing range reported here represent one effective configuration and performance can be further tailored to specific applications by adjusting the discretization parameters.

For each dynamic obstacle, we encode its position, velocity, and radius into a structured state vector. To ensure a fixed-length input compatible with the transformer architecture, we limit the maximum number of surrounding dynamic obstacles. Unlike static obstacles, dynamic obstacles require temporal information to capture motion patterns. Therefore, we incorporate a 2-second history with a sampling interval of 0.5 seconds, allowing the policy to infer short-term motion trends. The resulting dynamic obstacle state is formulated as:
\begin{equation}
\mathcal{S}_{\text{dyn}} = [ \mathcal{S}_{d,1}, \ldots, \mathcal{S}_{d,5}], \ \ \mathcal{S}_{d,i} = [ \mathbf{o}_{i}^{t-K}, \ldots, \mathbf{o}_{i}^{t}],
\end{equation}
where the state of the $i$-th dynamic obstacle at time $t$ is $\mathbf{o}_{i}^{t}=[\mathbf{p}_{i}^{t}, \mathbf{v}_{i}^{t}, r_i]$ with $\mathbf{p}_{i}^{t}$ denoting position, $\mathbf{v}_{i}^{t}$ velocity, and $r_i$ the obstacle radius. $K$ represents the length of the temporal history and $N$ denotes the maximum number of dynamic obstacles considered (set to 5 in our implementation). Although $N$ can be increased for highly crowded settings, five obstacles provide a practical balance between representational capacity and computational efficiency in local navigation.

Besides obstacle information, robot internal states are essential for goal-directed navigation and consist of the robot position $\mathbf{p}_r$, velocity $\mathbf{v}_r$, and the user-defined goal position $\mathbf{p}_g$.  To eliminate dependency on the global coordinate definition,  all vectors are expressed in a goal-aligned coordinate frame, where the $x$-axis is defined by the direction from the start position $\mathbf{p}_s$ to the goal $\mathbf{p}_g$, and the $z$-axis corresponds to the vertical direction. The internal state is defined as:
\begin{equation} \label{eqn: state input}
    \mathcal{S}_{\text{in}} = [\mathcal{S}^{t-K}, \ldots, \mathcal{S}^{t}], \mathcal{S}^{t} = [\frac{\mathbf{p}^{t}_{g} - \mathbf{p}^{t}_{r}}{\lVert \mathbf{p}^{t}_{g} - \mathbf{p}^{t}_{r} \rVert}, \lVert \mathbf{p}^{t}_{g} - \mathbf{p}^{t}_{r} \rVert, \mathbf{v}^{t}_{r}].
\end{equation} 
At each time step, the state encodes the unit direction vector toward the goal, the goal distance, and the robot velocity, all represented in the goal-aligned frame. Similar to the dynamic obstacle representation, we maintain a temporal history of $K$ steps corresponding to two seconds with a 0.5-second interval.

\textbf{Output Action Parameterization.} The action representation should be applicable across different robotic platforms. Therefore, we adopt velocity as the control output. For robots operating in 3D, such as UAVs, a 3D velocity vector is used, whereas for robots operating on a planar surface, such as quadruped robots, a 2D velocity vector is applied. The policy outputs a normalized velocity command $\hat{\mathbf{v}} \in [0,1]^d$ in the goal-aligned coordinate frame, where $d \in \{2,3\}$ depending on the platform. Given a predefined maximum velocity $v_{\max}$, the executable velocity action is computed as: 
\begin{equation}
\mathcal{A}_{\text{vel}} = [v^{x}, v^{y}, v^{z}], \  v^{i} = v_{\max} \left( 2 \hat{v}^{i} - 1 \right), \ i \in  {x,y,z}, 
\end{equation}
where $\hat{v}$ denotes the normalized velocity command for a single axis. To generate the bounded control, the network outputs the parameters $\alpha$ and $\beta$ of a Beta distribution. The probability density function of the Beta distribution is given by:
\begin{equation} \mathrm{Beta}(\hat{v};\alpha,\beta) = \frac{\Gamma(\alpha+\beta)}{\Gamma(\alpha)\Gamma(\beta)} \, \hat{v}^{\alpha-1} (1-\hat{v})^{\beta-1}, \hat{v} \in [0,1], \end{equation}
where $\Gamma(\cdot)$ denotes the Gamma function. Example probability density functions are illustrated in Fig. \ref{fig:formulation_representation}b. During training, actions are sampled from the Beta distribution, while its mean is used during deployment. Compared to Gaussian policies, Beta-distribution-based policies avoid boundary bias and typically achieve faster convergence under bounded action spaces \cite{chou2017improving}. Although some mobile robots employ alternative control such as linear speed and yaw rate, the proposed framework can be adapted by redefining the action parameterization.

\textbf{Reward Function Design.} The reward function is designed to guide the learning of navigation behaviors and is computed at each time step as a weighted sum of multiple components: 
\begin{equation}
    \mathcal{R} = \lambda_{1}\mathcal{R}_{v} + \lambda_{2}\mathcal{R}_{s} + \lambda_{3}\mathcal{R}_{sc} + \lambda_{4}\mathcal{R}_{dc} + \lambda_{6}\mathcal{R}_{sm} + \lambda_{6}\mathcal{R}_{h},
\end{equation}
where $\mathcal{R}_{(\cdot)}$ denotes an individual reward weighted by $\lambda_{i}$, and its detailed formulation is provided in the following.

\textit{a) Velocity reward $\mathcal{R}_{vel}$}: The velocity reward encourages the robot to generate velocity commands toward the goal.
\begin{equation}
    \mathcal{R}_{v} = \frac{\mathbf{p}_{g} - \mathbf{p}_{r}}{\lVert \mathbf{p}_{g} - \mathbf{p}_{r} \rVert} \cdot \mathbf{v}_{r}, \ \  \lambda_{1} = 1.0.
\end{equation}
Intuitively, velocity commands that align closely with the goal direction and exhibit higher speed are assigned larger rewards.

\textit{b) Survival reward $\mathcal{R}_{sc}$}: Since each episode is bounded by a maximum time horizon, the survival reward is introduced to incentivize continued operation in the environment and to promote long-horizon stability, defined as follows:
\begin{equation}
    \mathcal{R}_{s} = 1.0, \ \ \lambda_{2} = 1.0.
\end{equation}

\textit{c) Static collision reward $\mathcal{R}_{sc}$}: The static collision reward enforces safe distances from static obstacles and is formulated based on the static obstacle states defined in Eqn. \ref{eqn: static states} as follows: 
\begin{equation}
    \mathcal{R}_{sc} = \frac{1}{N_{h}N_{v}}\sum_{i=1}^{N_h}\sum_{j=1}^{N_v} \log\mathcal{S}_{\text{static}}(i, j), \ \lambda_{3} = 2.0, 
\end{equation}
where $N_{h}$ and $N_{v}$ denote the numbers of horizontal and vertical beams, respectively. The formulation computes the logarithm of the distance, leading to higher rewards when maintaining greater clearance from obstacles, while strongly penalizing behaviors that result in close proximity to obstacles.

\textit{d) Dynamic collision reward $\mathcal{R}_{dc}$}: Similar to the static collision reward, the dynamic collision reward is formulated based on the relative positions of dynamic obstacles:
\begin{equation}
    \mathcal{R}_{dc} = \frac{1}{N_{d}}\sum_{i=1}^{N_{d}} \log \lVert \mathbf{p}_{r} - \mathbf{p}_{o_{i}}\rVert, \ \lambda_{4} = 2.0, 
\end{equation}
where $N_{d}$ denotes the number of dynamic obstacles, and $p_{o_{i}}$ represents the position of the $i$-th dynamic obstacle. 

\textit{e) Smoothness reward $\mathcal{R}_{s}$}: The smoothness reward promotes temporally consistent control commands, leading to more natural and stable navigation behavior. It is defined as: 
\begin{equation}
    \mathcal{R}_{sm} = -\lVert \mathbf{v}^{t}_{r} - \mathbf{v}^{t-1}_{r}\rVert, \ \lambda_{5} = 0.1,
\end{equation}
where the $\mathcal{L}2$-norm of the difference between the robot velocities at consecutive time steps is penalized.

\textit{f) Height reward $\mathcal{R}_{h}$}: The height reward is applied only in 3D navigation scenarios. It is introduced to prevent the robot from trivially avoiding obstacles by excessively increasing its altitude. The reward is defined as the following:
\begin{equation} \mathcal{R}_{h} = -\min \left( \left| p_{r}^{z} - p_{s}^{z} \right|, \left| p_{r}^{z} - p_{g}^{z} \right| \right)^{2}, \  \lambda_{6} = 8.0, \end{equation}
where $p^{z}_{r}$, $p^{z}_{s}$ and $p^{g}_{r}$ denote the vertical positions of the robot, start point, and goal point, respectively. This term penalizes deviations from the height range defined by the start and goal positions, discouraging unnecessary altitude changes.

\subsection{Policy Network Design for Temporal Reasoning} \label{sec:network_design}
This section describes the neural network architecture for navigation policy. The core of the policy network is a transformer-based backbone designed to integrate static obstacles, dynamic obstacles, and robot internal states over time.

\textit{a) Token Design and Temporal Encoding:} The transformer operates on a fixed-length sequence of 12 tokens for computational efficiency. One token is reserved as a \texttt{[CLS]} token to aggregate global information and serve as the feature representation for downstream policy and value prediction. Temporal information is incorporated by allocating five tokens for dynamic obstacle history and five tokens for robot internal state history. A single token encodes static obstacle information, which is time-invariant within each step. Positional encodings are added to preserve temporal ordering.

\textit{b) Modality Alignment:} To align heterogeneous modalities within a unified token embedding space, the static obstacle distance states are first processed by a lightweight convolutional neural network encoder consisting of three layers with channel sizes of 4, 16, and 16, followed by flattening to extract spatial features. The resulting static obstacle features, together with dynamic obstacle states and robot internal states, are then encoded and projected into a shared embedding dimension using a token projection MLP with layer sizes [128, 64].

\textit{c) Transformer Backbone:} The transformer backbone has an embedding dimension of 64, four attention heads, and four encoder layers, with a feedforward dimension of 128 and dropout rate of 0.1. Through self-attention, the network jointly integrates static obstacle states, robot internal state histories, and dynamic obstacle histories, enabling implicit modeling of temporal dependencies and interaction dynamics.

\textit{d) Actor–Critic Heads:} The aggregated \texttt{[CLS]} feature from the transformer backbone is fed into separate MLP-based actor and critic networks, each composed of two hidden layers with 256 neurons. The actor predicts the parameters of the action distribution, while the critic estimates the state value.

The entire policy neural network contains 0.31M parameters, which enables efficient real-time deployment.
\begin{figure}[t] 
    \centering
    \includegraphics[scale=1.15]{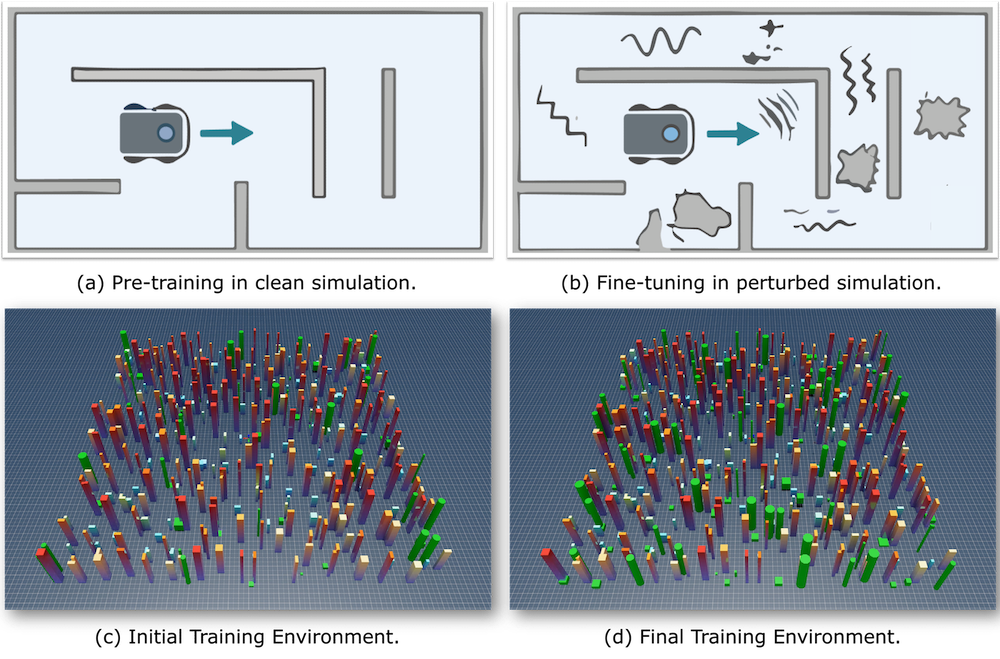}
    \caption{Illustration of the proposed training strategy. (a) The policy is initially trained in clean simulation settings and (b) subsequently fine-tuned under controlled perturbations to improve generalization and robustness. (c) Curriculum learning is applied by starting with low obstacle density and (d) progressively increasing the environment complexity to high-density settings.}
    \label{fig:training_strategy}
\end{figure}

\subsection{RL Training Strategy} \label{sec: training strategy}
This section presents the overall training methodology. As illustrated in Fig. \ref{fig:training_strategy}, we adopt a two-stage training strategy consisting of a pre-training phase in clean simulation environments, followed by a post-training phase that enables the policy to adapt to a broader range of conditions.

\textbf{Pre-training.} The pre-training stage enables the policy to acquire fundamental navigation behaviors in complex yet clean simulation environments without additional perturbations. We conduct training using a quadcopter UAV platform that adopts velocity commands as high-level control inputs, which are executed through a low-level flight controller. To accelerate convergence and improve sample efficiency, training is implemented in NVIDIA Isaac Sim, which supports massively parallel simulation. In our implementation, 1024 robots are executed in parallel for data collection. We employ Proximal Policy Optimization with Generalized Advantage Estimation (GAE) \cite{schulman2015high} for stable and efficient policy learning. All networks are optimized using the Adam optimizer with a learning rate of $2 \times 10^{-4}$. To encourage sufficient exploration during training, an entropy regularization term is incorporated into the objective function. The overall loss is formulated as:
\begin{equation}
    \mathcal{L}_{\text{total}} = \mathcal{L}_{\text{actor}} + \mathcal{L}_{\text{critic}} + \mathcal{L}_{\text{entropy}}.
\end{equation}
The training environment (Fig. \ref{fig:training_strategy}c-d) contains randomly distributed box- and cylinder-shaped 
static obstacles with varying sizes. Dynamic obstacles move with randomly 
sampled speeds in the range $[0, 1.5]$~m/s and change their motion direction 
every $s$ seconds, where $s$ is uniformly sampled from $[0, 2]$.

\textbf{Post-training.} We adopt perturbation-aware fine-tuning as a post-training adaptation strategy to improve robustness and enhance sim-to-real transfer performance. Based on our prior deployment experience, we identify several key factors that affect real-world performance. We observe that explicitly incorporating perturbations in these factors during fine-tuning substantially improves policy robustness under sim-to-real discrepancies. The specific factors are detailed below.

\textit{a) Input Noise:} The most straightforward source of sim-to-real discrepancy arises from input noise. Based on the state definitions in Eqn. \ref{eqn: static states}-\ref{eqn: state input}, the primary noise sources include static obstacle distance estimation, dynamic obstacle state estimation, and the robot’s own state estimation. These inaccuracies stem from both sensor noise and the algorithms used for state estimation. To improve robustness to these estimation errors, we inject zero-mean Gaussian noise into the input states during training, with empirically determined standard deviations.

\textit{b) Perception Failure:} Dynamic obstacle perception is prone to failure, most commonly due to false negatives, which prevent the robot from detecting moving objects that may pose collision risks. To simulate such perception failures during training, we randomly drop detected obstacles using Bernoulli sampling with a predefined probability. In our implementation, the dropout probability is set to $p_{\text{drop}} = 0.3$ based on a conservative estimate of the empirical perception failure rate.

\textit{c) Input Latency:} Latency is often overlooked in sim-to-real transfer, yet it can have a non-negligible impact on deployment performance. Input latency arises from multiple sources, including occupancy map construction, dynamic obstacle detection and tracking, 
robot state estimation, and data communication. Although these modules are designed to operate at high frequency, they can still introduce delays on the order of several to tens of milliseconds. To account for this effect, we explicitly model input latency during training by introducing random delays sampled from a uniform distribution. In our implementation, the delay is sampled from the interval $[0, 0.1]$ seconds, based on experimental measurements.

\textit{d) Action Latency:} Similar to input latency, action execution can also be delayed due to network computation overhead and data communication. To model this effect, we introduce random delays to the action output during training, with delays sampled uniformly from the interval $[0, 0.04]$ seconds based on empirical estimates of data communication latency.

\textit{e) Control Response:} To ensure that the learned policy can generalize across robots with different low-level controllers and control dynamics, we introduce variability in control response during training. Specifically, we randomize the velocity controller gain within $\pm 40\%$ of its nominal value (from 60\% to 140\%), covering a wide range of mobile robotic platforms, including UAVs with different masses and quadruped robots. This range corresponds to rise times from 1.12\,s to 0.32\,s and settling times from 2.14\,s to 0.96\,s, respectively.

During the post-training stage, we incorporate the previously described perturbations into the simulation and fine-tune the policy with a reduced learning rate of $5 \times 10^{-5}$.

\textbf{Curriculum Learning.} Rather than training directly in the most challenging environment, we adopt a curriculum learning strategy that progressively increases task difficulty. This is implemented by gradually increasing obstacle density during training. We observe that curriculum learning significantly improves the final policy performance. The curriculum stages are summarized in Table \ref{tab:curriculum learning}, with obstacle density levels ranging from S1 to S5. Examples of the initial and final training environments are shown in Fig. \ref{fig:training_strategy}c-d. During pre-training, stages S1 to S4 are applied, while post-training uses stages S1 to S3. Note that this curriculum stage design is guided by empirical evaluation results, as discussed in Sec. \ref{sec: result}.
\begin{table}[t]
\centering
\setlength{\tabcolsep}{6pt}
\renewcommand{\arraystretch}{1.22}
\caption{\textbf{Curriculum Learning Stages.} 
Each stage specifies the numbers of static and dynamic obstacles during training.}
\begin{tabular}{c l}
\toprule
\textbf{Stage} & \textbf{Training Environment Configuration} \\
\midrule
S1 & 300 static + 60 dynamic obstacles \\
S2 & 350 static + 80 dynamic obstacles \\
S3 & 400 static + 100 dynamic obstacles \\
S4 & 400 static + 120 dynamic obstacles \\
S5 & 400 static + 140 dynamic obstacles \\
\bottomrule
\end{tabular}
\label{tab:curriculum learning}
\end{table}

\subsection{Real-World Deployment} \label{sec:deployment}
With the trained policy, this section describes the real-world deployment strategy. We first introduce the perception module used to process raw sensor data, followed by the safety shield mechanism designed to prevent potential collisions. To demonstrate the practical applicability of the learned policy, 
we present its integration into navigation-centric tasks, 
including unknown environment exploration and autonomous inspection.

\textbf{Perception Module.} The perception module consists of occupancy mapping for static structures and detection and tracking for dynamic obstacles. For static obstacles, we adopt a standard occupancy formulation with the recursive update:
\begin{equation}
\ell_t(x) = \ell_{t-1}(x) + \log \frac{p(x \mid z_t)}{1 - p(x \mid z_t)} - \log \frac{p(x)}{1 - p(x)},
\label{eq:occupancy_update}
\end{equation}
where $\ell_t(x)$ denotes the log-odds of voxel $x$ being occupied at time $t$, and $z_t$ represents the sensor observation obtained from a depth camera or LiDAR point cloud. A fixed-size voxel map is predefined according to the operating environment, with occupancy values stored in an indexed array for efficient access ($\mathcal{O}(1)$ time complexity). Although this recursive update scheme is computationally efficient and well suited for modeling static structures, its temporal accumulation property makes it less suitable for rapidly moving dynamic obstacles.

\begin{figure}[t] 
    \centering
    \includegraphics[scale=0.53]{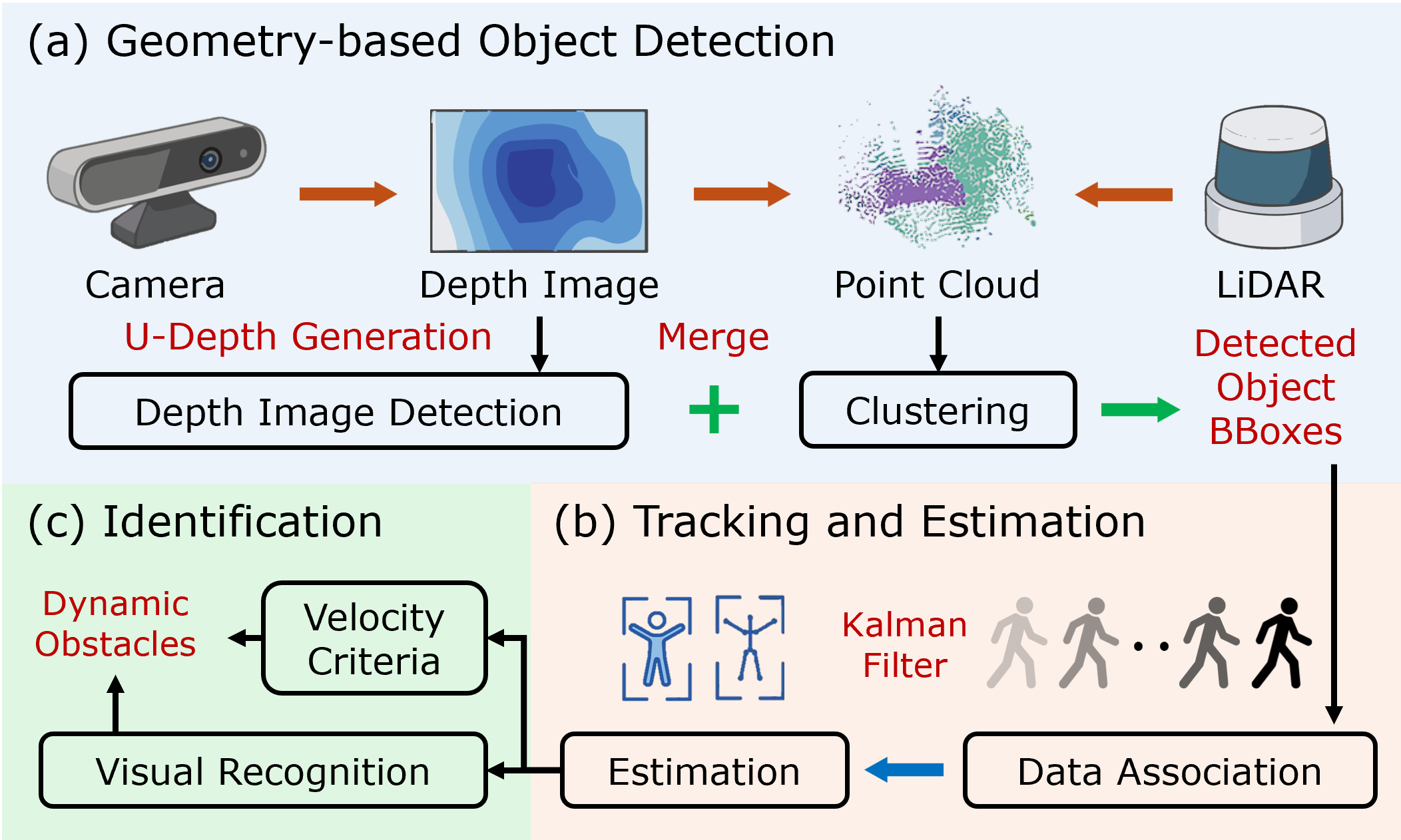}
    \caption{Illustration of the perception module. (a) Raw sensor data, including depth images or point clouds, are processed to generate detected object bounding boxes. (b) Detected objects are temporally associated to estimate their motion. (c) Dynamic obstacles are identified based on predefined criteria. }
    \label{fig:percetion_module}
\end{figure}

The dynamic obstacle perception pipeline is illustrated in Fig. \ref{fig:percetion_module}. At each perception cycle, incoming raw sensor data are processed to detect obstacles based on geometric cues, representing both static and dynamic objects as axis-aligned 3D bounding boxes without orientation estimation (Fig. \ref{fig:percetion_module}a). Since different sensors produce different data characteristics, detection strategies are modality-dependent. For LiDAR point clouds, which offer wide-range and geometrically accurate measurements, we first apply downsampling and then perform DBSCAN clustering to extract point clusters and estimate their centroids and spatial extents, yielding 3D bounding boxes. In contrast, RGB-D depth data are denser but noisier, making clustering alone insufficient for robust detection. To address this, we additionally employ a U-depth map–based method \cite{lin2020robust} that leverages line grouping in the image domain to generate candidate bounding boxes. The outputs of the depth clustering and U-depth detection are combined through an ensemble agreement strategy \cite{detection} to improve accuracy. For robotic platforms equipped with both LiDAR and RGB-D camera sensors, detection is performed independently for each sensor data and the results are subsequently combined.

The second stage estimates dynamic obstacle states using temporal information, as illustrated in Fig. \ref{fig:percetion_module}b. We first perform data association to match newly detected obstacles with those observed in previous time steps. For each previously tracked obstacle, the current position is predicted from its historical state using linear propagation under a constant-velocity assumption. Matching is then performed based on a cosine similarity score between the feature vectors of predicted and detected states. The velocity of each newly detected obstacle is initially estimated from the ratio of position change to the corresponding time interval. The feature vector is defined as: 
\begin{equation}
    \phi^{\text{feat}}_{i} = \left[ \mathbf{p}_{i}, \mathbf{v}_{i}, r_i \right], 
    \  \mathbf{p}_{i} \in \mathbb{R}^2, \mathbf{v}_{i} \in \mathbb{R}^2,
\end{equation}
where $\mathbf{p}_{i}$ and $\mathbf{v}_{i}$ denote the obstacle position and velocity in the horizontal plane, and $r_i$ represents its radius.  We assume that dynamic obstacles move only within the planar space. After data association, a Kalman filter is applied to refine and update the obstacle states over time, producing consistent position and velocity estimates for all tracked obstacles.

The final stage of dynamic obstacle perception is to identify moving objects among all tracked obstacles, as illustrated in Fig. \ref{fig:percetion_module}c. This is achieved through two complementary criteria. First, an obstacle is classified as dynamic if its estimated velocity exceeds a predefined threshold. Second, if the obstacle lies within the camera field of view and is recognized by a 2D deep learning–based object detector as belonging to a predefined dynamic category, it is also labeled as dynamic. 

\textbf{Policy Safety Shield.} To ensure safety during deployment, the policy output is optionally processed by a safety shield that filters unsafe commands and projects them into a safe velocity set based on the classical Velocity Obstacle (VO) formulation \cite{velocity_obstacle}. The velocity obstacle defines the set of robot velocities that would lead to a collision with an obstacle within a specified future time horizon. Given the policy output velocity $\mathbf{v}_{\pi}$ and the velocity of the $i$-th obstacle $\mathbf{v}_{i}$, we construct the corresponding velocity obstacle in the relative velocity space $\mathbf{v}_{\pi} - \mathbf{v}_{i}$. For static obstacles, the distance-based representation is converted into a position–radius representation with zero obstacle velocity. If the policy velocity lies either inside or outside the velocity obstacle region, we compute the minimum velocity adjustment $\delta \mathbf{v}_{i}$ that moves the relative velocity to the closest point on the VO boundary. The geometric computation of $\delta \mathbf{v}_{i}$ follows the standard construction in \cite{velocity_obstacle}. The safe velocity is obtained by solving the following projection problem:
\begin{mini!}[2]
    {\mathbf{v}_{safe}}{\begin{Vmatrix} \mathbf{v}_{safe} - \mathbf{v}_{\pi} \end{Vmatrix}_{2}^{2},}{}{} 
\addConstraint{(\mathbf{v}_{safe} - (\mathbf{v}_{\pi} + \delta \mathbf{v}_{i})) \cdot \delta \mathbf{v}_{i}}  \geq 0{}{} \label{plane constraint}
\addConstraint{\mathbf{v}_{min} \leq \mathbf{v}_{safe} \leq \mathbf{v}_{max}}{} \label{control limits}
\addConstraint{\forall i \in \{1, \ldots , N\}},
\end{mini!}
where the objective minimizes the deviation from the policy output, and the first constraint defines a half-space that ensures the projected velocity lies outside the velocity obstacle region for each obstacle. The second constraint enforces control limits. This safety projection prevents collision-inducing commands that may arise from the black-box nature of neural network output actions. However, due to the conservative nature of the VO formulation, a feasible solution may not exist in highly cluttered environments, in which case the robot defaults to zero velocity. Therefore, the safety shield is designed as an optional module, allowing a trade-off between safety and performance depending on the deployment environments.

\begin{algorithm}[t] \label{alg: application}
\caption{Navigation-Centric Task Deployment} 
\SetAlgoNoLine%
$C_{\text{complete}} \gets False$\; \label{complete_flag}
$\text{PM} \gets \text{Perception Module}$\; \label{perception_module_alg}
$\xi^{r} \gets \text{Current Robot States}$\; \label{state_estimation_module_alg}
$\mathcal{P}_{g} \gets \text{Global Task Planner}$ \Comment*[r]{Inspect/Explore}
$\mathcal{P}_{rl} \gets \text{RL Navigation Policy}$\; \label{RL_policy}
\While{\normalfont{\textbf{not}} $C_{\text{complete}}$}{ \label{task_while_loop}
    $\mathcal{M} \gets \text{PM}.\textbf{getUpdatedMap}(\xi^{r})$\;
    $p_{\text{goal}} \gets \mathcal{P}.\textbf{getTaskGoalPosition}(\mathcal{M}, \xi^{r})$ \label{goal_planning}

    $C_{\text{reach}} \gets False$\; \label{low_level_start}
    \While{\normalfont{\textbf{not}} Reach}{
        $\mathcal{M} \gets \text{PM}.\textbf{getUpdatedMap}(\xi^{r})$\;
        $\mathcal{D} \gets \text{PM}.\textbf{getDynamicObstacles}(\xi^{r})$\;
        $\mathcal{A}_{output} \gets \mathcal{P}_{rl}.\textbf{getAction}(\mathcal{M}, \mathcal{D}, \xi^{r})$\; \label{low_level_end}
    }
}
\end{algorithm}
\textbf{Task Deployment.} Most navigation-centric applications are traditionally implemented using model-based or optimization-based methods. Beyond proposing a reinforcement learning–based navigation framework, we further demonstrate its applicability to real-world tasks. Specifically, we evaluate two representative applications: autonomous exploration in unknown environments and autonomous surface inspection. In the exploration task, the robot is deployed without a prior map and must actively determine successive goal locations to build a map of the environment. In contrast, surface inspection assumes a known map and generates a sequence of waypoints that the robot must traverse. Despite their differences, both tasks rely on accurate waypoint tracking and robust collision avoidance in the presence of static and dynamic obstacles.

To integrate the RL navigation policy into these applications, we follow the pipeline illustrated in Alg. \ref{alg: application}. With the completion flag initialized to false (line \ref{complete_flag}), the perception and state estimation modules are initialized (Lines \ref{perception_module_alg}–\ref{state_estimation_module_alg}), together with the global task planner and the RL navigation policy (Line \ref{RL_policy}). At each planning iteration, while the task remains incomplete, the global planner updates the environment representation and generates the next goal position (Lines \ref{task_while_loop}–\ref{goal_planning}). The goal is then passed to the RL navigation policy, which produces control actions based on updated map and dynamic obstacle information (Lines \ref{low_level_start}-\ref{low_level_end}). This process repeats until the robot reaches the designated goal. Although demonstrated on exploration and inspection tasks, the proposed framework is not limited to these scenarios and can be readily extended to other mobile robot applications that require waypoint-based navigation with real-time collision avoidance.

\section{Result and Discussion} \label{sec: result}
This section presents the experimental evaluation of the proposed framework. We consider three types of evaluation. First, we conduct standard simulation evaluation, where the trained policy is tested in the same simulator used for training. To better approximate real-world deployment, we introduce the perturbations described in Sec. \ref{sec: training strategy} during evaluation. Second, we evaluate cross-simulator transfer performance, where the trained policy is deployed in a different simulator with distinct physics engines, sensor models, and robot parameters (including mass and dynamics). This setting assesses the robustness of the policy to discrepancies in dynamics and perception models that were not encountered during training. Finally, we conduct extensive real-world experiments on both aerial and legged robotic platforms using the same trained policy. This demonstrates the generalization capability of the proposed framework across different robot embodiments and previously unseen environments. In addition, we validate the policy on navigation-centric tasks to further demonstrate its practical applicability in real-world scenarios. Specifically, this section aims to answer the following questions:
\begin{itemize}
\item \textit{What is the quantitative performance of the framework when evaluated in the same training environments with perturbations, and how does each proposed component contribute to the overall performance?} Sec. \ref{sec:intra-sim-results}.
\item \textit{How do the introduced perturbations affect performance quantitatively, and how do different deployment velocities and control frequencies influence the results?} Sec. \ref{sec:sim-to-real pertubations}.
\item \textit{How does the proposed method perform when transferred to a different simulator (cross-simulator transfer performance), and how does it compare with model-based planners in complex environments?} Sec. \ref{sec:cross-sim-results}.
\item \textit{What are the key parameters that influence the training performance of RL-based navigation?} Sec. \ref{sec:training-parameters}
\item \textit{How does the policy perform in real-world deployment qualitatively, and can it reliably accomplish navigation-centric tasks using the learned RL policy?} Sec. \ref{sec:real-world-results}.
\end{itemize}

\begin{figure}[t] 
    \centering
    \includegraphics[scale=0.55]{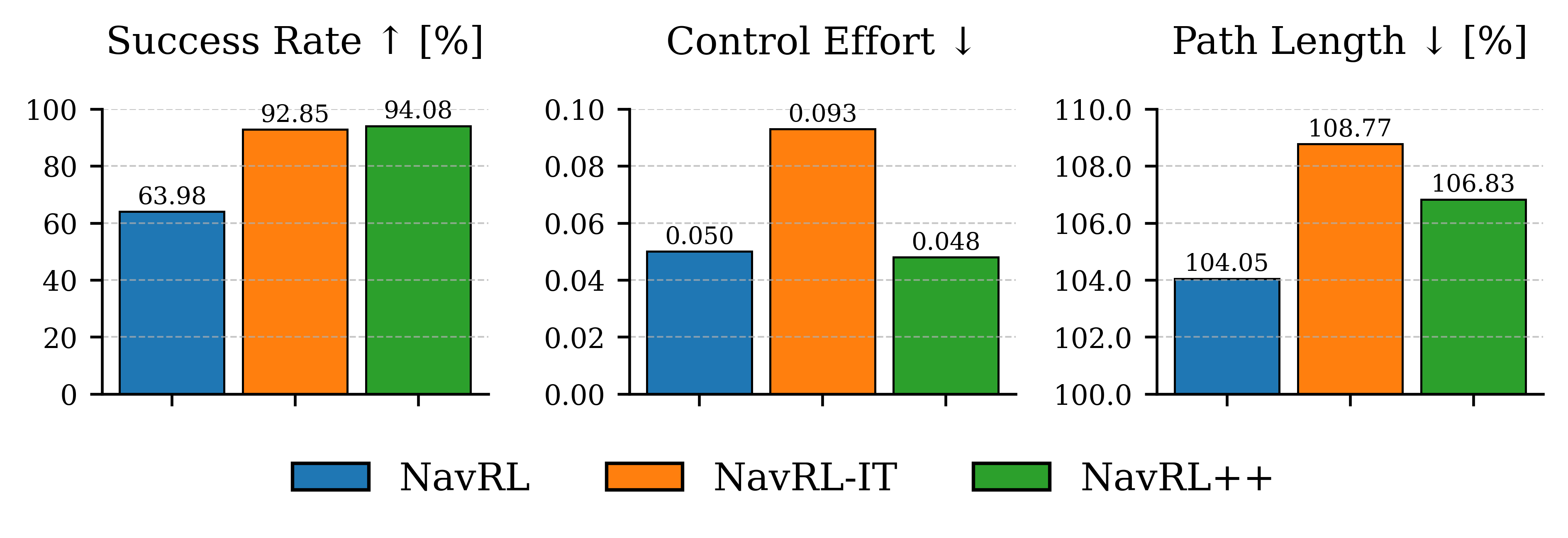}
    \caption{Overall performance benchmarking among baseline RL approaches.}
    \label{fig:overall_metrics}
\end{figure}

\subsection{Implementation Details \& Metrics} 

\begin{figure}[t] 
    \centering
    \includegraphics[scale=0.57]{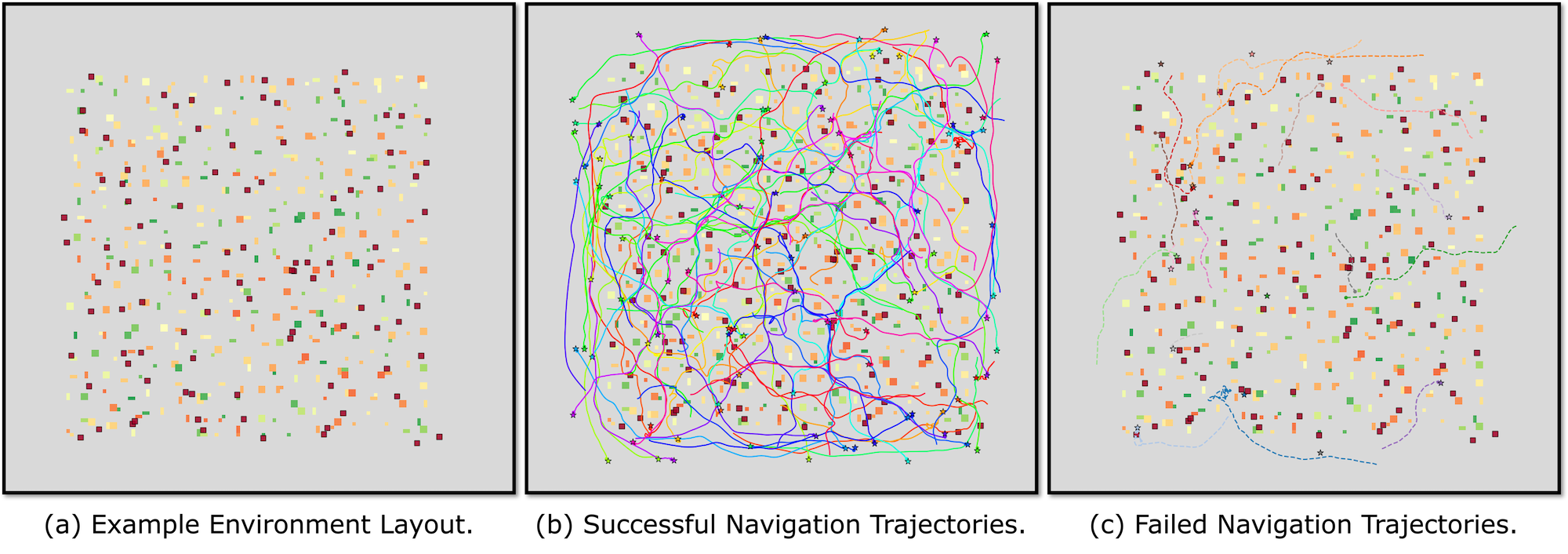}
    \caption{Visualization of simulation evaluation with 100 (out of 10,000) randomized start–goal trials in an environment containing 400 static obstacles and 120 dynamic obstacles. (a) Example environment, where static obstacles are shown as colored boxes and dynamic obstacles as red boxes. (b) Eighty-four successful navigation trajectories. (c) Sixteen failed trajectories. }
    \label{fig:experiment_examples}
\end{figure}

\begin{table*}[t]
\centering
\setlength{\tabcolsep}{3.6pt}
\renewcommand{\arraystretch}{1.2}
\caption{Comparison of training methods across varying environment complexity levels under static and dynamic environments.}
\begin{tabular}{lcccccccccccccc}
\toprule
\multirow{4}{*}{\parbox{1.5cm}{\raggedright\textbf{Evaluation Method}}}
  & \multicolumn{12}{c}{\textbf{Intra-Sim2Sim Transfer under Domain Perturbation}}
  & \multicolumn{2}{c}{\textbf{Overall}} \\
\cmidrule(lr){2-13}\cmidrule(lr){14-15}
  & \multicolumn{4}{c}{Low-Complexity}
  & \multicolumn{4}{c}{Mid-Complexity}
  & \multicolumn{4}{c}{High-Complexity}
  & \multicolumn{2}{c}{All-Scenarios} \\
\cmidrule(lr){2-5}
\cmidrule(lr){6-9}
\cmidrule(lr){10-13}
\cmidrule(lr){14-15}
  & \multicolumn{2}{c}{Static}
  & \multicolumn{2}{c}{Dynamic}
  & \multicolumn{2}{c}{Static}
  & \multicolumn{2}{c}{Dynamic}
  & \multicolumn{2}{c}{Static}
  & \multicolumn{2}{c}{Dynamic}
  & \multicolumn{2}{c}{Combined} \\
  & SR [\%] $\uparrow$ & CE $\downarrow$
  & SR [\%] $\uparrow$ & CE $\downarrow$
  & SR [\%] $\uparrow$ & CE $\downarrow$
  & SR [\%] $\uparrow$ & CE $\downarrow$
  & SR [\%] $\uparrow$ & CE $\downarrow$
  & SR [\%] $\uparrow$ & CE $\downarrow$
  & SR [\%] $\uparrow$ & CE $\downarrow$\\
\midrule

NavRL \cite{NavRL}
& 85.53 & \textbf{0.037}
& 52.16 & 0.061
& 81.45 & \textbf{0.038}
& 43.94 & 0.061
& 76.86 & \textbf{0.039}
& 38.35 & 0.061
& 63.05 & 0.050 \\

NavRL-IT
& 99.59 & 0.084
& 90.25 & 0.099
& 99.40 & 0.085
& 86.28 & 0.102
& 98.79 & 0.087
& 82.79 & 0.101
& 92.85 & 0.093 \\

NavRL-IT-T
& 99.81 & 0.043
& 86.45 & \textbf{0.042}
& 99.79 & 0.043
& 81.09 & \textbf{0.042}
& 99.60 & 0.044
& 76.49 & \textbf{0.042}
& 90.54 & \textbf{0.043} \\

NavRL-IT-PF
& 66.42 & 0.058
& 88.97 & 0.099
& 68.86 & 0.062
& 85.75 & 0.090
& 71.00 & 0.087
& 82.80 & 0.091
& 77.30 & 0.081 \\

\rowcolor{gray!20}
\textbf{NavRL++ (Ours)}
& \textbf{99.90} & 0.046
& \textbf{92.49} & 0.051
& \textbf{99.94} & 0.047
& \textbf{88.32} & 0.050
& \textbf{99.84} & 0.047
& \textbf{83.96} & 0.049
& \textbf{94.08} & 0.048 \\

\bottomrule
\end{tabular}
\label{tab:benchmark_results}
\end{table*}

\begin{table}[t]
\centering
\setlength{\tabcolsep}{5pt}
\renewcommand{\arraystretch}{1.2}
\caption{Intra-simulator evaluation of path length in static long-distance (48 $m$) navigation scenarios.}
\begin{tabular}{lccccc}
\toprule
\textbf{Path Length [m] $\downarrow$} & \textbf{Low} & \textbf{Mid} & \textbf{High} & \textbf{Average} & \textbf{Percentage} \\
\midrule
NavRL \cite{NavRL}        
& \textbf{49.66} 
& \textbf{49.99}  
& \textbf{50.16}  
& \textbf{49.94}  
& \textbf{104.05\%} \\

NavRL-IT       
& 51.92  
& 52.13  
& 52.59  
& 52.21  
& 108.77\% \\

NavRL-IT-PF    
& 50.77  
& 50.94  
& 50.85  
& 50.85  
& 105.94\% \\

NavRL-IT-T 
& 50.71  
& 50.63  
& 50.84  
& 50.73  
& 105.69\% \\

\rowcolor{gray!20}
\textbf{NavRL++ (Ours)}
& 51.16  
& 51.21  
& 51.47  
& 51.28  
& 106.83\% \\
\bottomrule
\end{tabular}

\label{tab:static_long_distance}
\end{table}

The reinforcement learning  policy is trained entirely in NVIDIA Isaac Sim using 1,024 robots running in parallel under a curriculum learning framework (Table \ref{tab:curriculum learning}). During the pre-training stage, the first curriculum level (S1) requires approximately 80 GPU hours on an NVIDIA RTX 4090. The subsequent curriculum stages (S2–S4) each require approximately 20–30 GPU hours. In the fine-tuning phase, each curriculum stage (S1–S3) requires an additional 10–15 GPU hours. In total, the complete training process consumes roughly 200 GPU hours, corresponding to more than 80,000 hours of aggregated experience due to parallel simulation.

For evaluation and deployment in simulation, the trained policy is implemented in both NVIDIA Isaac Sim and Gazebo. Isaac Sim is used for intra-simulator performance evaluation and ablation studies. Gazebo, in contrast, serves as a cross-simulator platform to evaluate transfer robustness under different physics and rendering settings. In addition, Gazebo enables benchmarking against representative optimization-based baseline planners.

The real-world deployment platforms are shown in Fig. \ref{fig:robot_platform}. We deploy the same trained policy (2D output) on two robotic platforms: a custom-built UAV and a Unitree quadruped robot. Both platforms share a similar sensor configuration, consisting of an Intel RealSense RGB-D camera and a Livox Mid-360 LiDAR. To evaluate sensor robustness, we test three perception configurations: camera-only, LiDAR-only, and camera–LiDAR fusion. The state estimation module is configured accordingly, using either vision-based \cite{vins} or LiDAR-based \cite{fast-lio2} state estimation depending on the selected sensing modality.

For quantitative evaluation, we consider three primary metrics. The most important metric is the success rate (SR), reported as a percentage, which measures the percentage of robots that successfully complete the navigation task within the allowed time. The second metric is the control effort (CE), measured in $m/s^2$. It is computed as the sum of the norms of velocity changes between consecutive control steps (with a sampling interval of $t=0.1$s), divided by the total navigation time. This metric reflects the smoothness and stability of the control actions. The third metric is the path length, measured in meters, representing the total traveled distance from the start to the goal and measuring navigation efficiency. 

For evaluation, we consider six environments. Three static environments contain randomly distributed cylindrical and box-shaped obstacles, with 300, 350, and 400 obstacles placed within a 40×40 m area. For the dynamic settings, we retain the same static obstacles and additionally introduce 60, 80, and 100 dynamic obstacles. The dynamic obstacles move at randomly sampled speeds within [0, 1.5] m/s and change their motion direction at random intervals between 0 and 2 seconds. These configurations correspond to low, medium, and high-density scenarios for both static and dynamic environments. Unless specified, the maximum velocity is set to 2 m/s.

\begin{table}[t]
\centering
\renewcommand{\arraystretch}{1.1}
\setlength{\tabcolsep}{7pt}
\caption{Comparison of navigation success rates under different output action limits and control frequencies.}
\label{tab:random_goal_vel_freq}

\begin{tabular}{p{1.0cm} c c c c}
\toprule
\textbf{Type} & \textbf{Setting} & \textbf{Static SR} & \textbf{Dynamic SR} & \textbf{Overall SR} \\
\midrule
Velocity & 1.5 m/s              & 99.77\% & 81.09\% & 90.43\% \\
& 2.0 m/s$^{\dagger}$  & 99.89\% & 88.26\% & 94.08\% \\
& 2.5 m/s              & 99.77\% & 89.53\% & 94.68\% \\
& 3.0 m/s              & 99.07\% & 87.42\% & 93.25\% \\
\midrule
Freq.
& 2 Hz                 & 96.47\% & 72.92\% & 84.69\% \\
& 5 Hz                 & 99.65\% & 82.48\% & 91.06\% \\
& 10 Hz                & 99.82\% & 86.58\% & 93.20\% \\
& 20 Hz                & 99.88\% & 88.18\% & 94.03\% \\
& 50 Hz$^{\dagger}$    & 99.89\% & 88.26\% & 94.08\% \\
\bottomrule
\end{tabular}

\vspace{2pt}
\footnotesize{$^{\dagger}$ Shared baseline configuration.}
\end{table}

\begin{figure*}[t]
    \centering
    \includegraphics[scale=1.23]{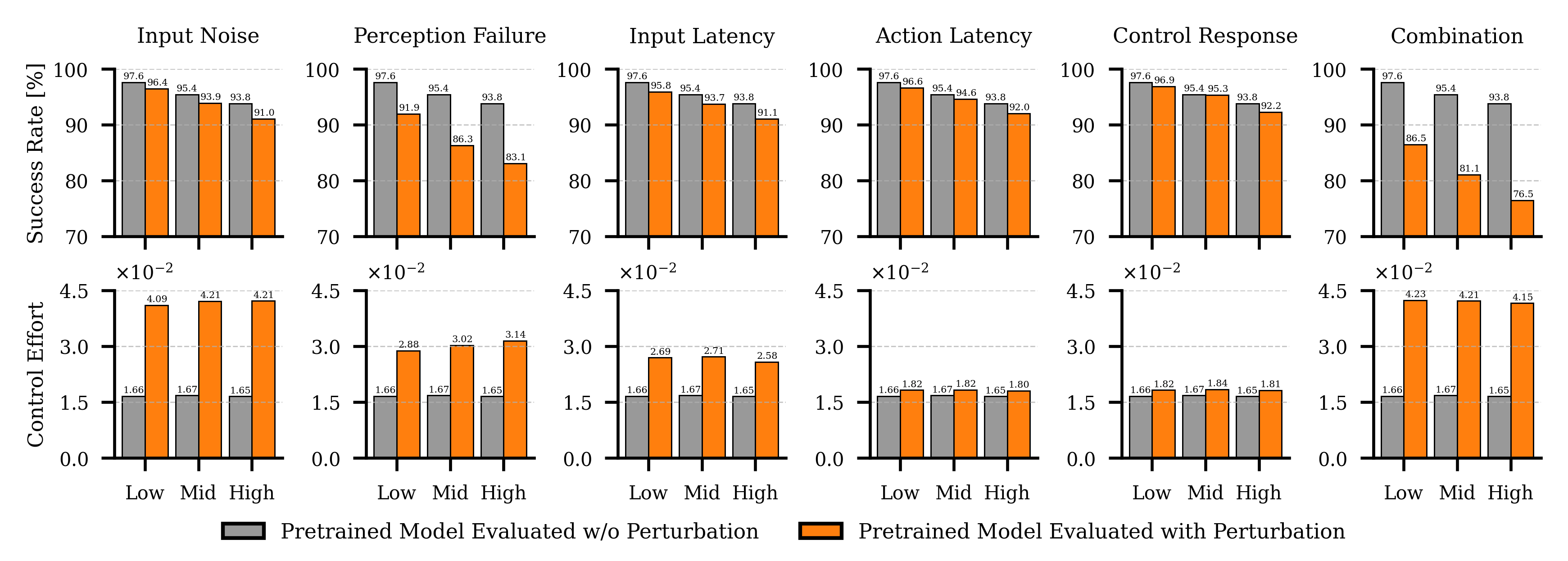}
    \caption{Effect of individual perturbations and their combinations on success rate and control effort of a policy trained in clean simulation.}
    \label{fig:perturbation_ablation}
\end{figure*}

\subsection{Performance Evaluation \& Ablation Studies}\label{sec:intra-sim-results}
This section presents a comprehensive evaluation of the proposed navigation framework in Isaac Sim environments that match the training environments while incorporating perturbations during evaluation. We further compare the proposed method with the baseline approach, \textit{NavRL}~\cite{NavRL}, to isolate the contribution of each proposed component to the overall RL-based navigation performance. As summarized in Table \ref{tab:benchmark_results}, we evaluate four variants in addition to our full model. \textit{NavRL} denotes the original baseline. \textit{NavRL-IT} denotes improved training strategies, including implementation refinements as well as parameter optimizations detailed in Sec. \ref{sec:training-parameters}. \textit{NavRL-IT-T} replaces the original single-step convolutional policy with our proposed temporal-aware network while retaining the improved training scheme. \textit{NavRL-IT-PF} augments the improved training with perturbation-aware fine-tuning using the convolutional architecture. Finally, our proposed method (\textit{NavRL++}) combines improved training, a transformer-based policy network, and perturbation-aware fine-tuning. Each data point is computed as the average over 10,000 navigation trials with randomly sampled start and goal positions. Fig. \ref{fig:experiment_examples} visualizes 100 example runs in a dynamic environment including both successful and failed cases and Fig. \ref{fig:isaac_sim_qualitative_experiments} visualize the experiments in Isaac Sim environments.

In Table \ref{tab:benchmark_results}, \textit{NavRL++} demonstrates improved deployment robustness across both static and dynamic environments while maintaining comparatively low control effort. Comparing \textit{NavRL-IT} with \textit{NavRL}, we observe that the improved training strategies substantially increase the final success rate, demonstrating their effectiveness. However, this improvement comes at the cost of significantly higher control effort, where pronounced action oscillations are observed. 

By replacing the single-step convolutional network with a transformer-based backbone, \textit{NavRL-IT-T} reduces the control effort from 0.093 to 0.043 $m/s^2$ while maintaining a comparable success rate, highlighting the importance of temporal reasoning in producing smoother actions. Furthermore, comparing \textit{NavRL-IT-PF} with \textit{NavRL-IT} shows that perturbation-aware fine-tuning does not improve the convolutional policy. In contrast, when applied to the transformer-based policy (\textit{NavRL++} vs. \textit{NavRL-IT-T}), perturbation-aware fine-tuning leads to clear performance gains, particularly in dynamic environments. This demonstrates the effectiveness of combining temporal reasoning with perturbation-aware adaptation.

The path length comparison is presented in Table \ref{tab:static_long_distance}. Unlike the previous experiments, all methods are evaluated on the same set of 10,000 fixed start–goal pairs with a 48-meter separation, and failure cases are excluded. The results show that all methods achieve similar path lengths (within 5\% variation). Although \textit{NavRL} yields slightly shorter paths, the differences are marginal. As visualized in Fig. \ref{fig:overall_metrics}, \textit{NavRL++} achieves the highest success rate with low control effort and short path length, demonstrating the best overall performance.

\subsection{Impact of Sim-to-Real Perturbations} \label{sec:sim-to-real pertubations}
The discrepancy between simulation and the real world is a primary factor limiting the robustness of RL-based navigation policies. To quantify the impact of sim-to-real perturbations, we evaluate a policy trained in a clean simulation setting without perturbation-aware fine-tuning (\textit{NavRL-IT-T}). The policy is tested under individual perturbations in dynamic environments with low, medium, and high-density obstacle environments. The results are summarized in Fig. \ref{fig:perturbation_ablation}. Overall, all introduced perturbations degrade performance. Perception failure has the largest impact on success rate, causing more than a 5\% drop across all difficulty levels. Other factor (input noise, latency, and control disturbance) lead to moderate reductions of approximately 1–3\%, depending on environment complexity. When combined, these perturbations produce a compounded effect, reducing the success rate by over 10\%, and by more than 15\% in high-density environments. Control effort follows a similar trend. Input noise introduces the most significant increase, while perception failure and input latency also noticeably degrade smoothness. Action latency and control response mismatch have smaller effects individually, but the combined perturbations increase control effort by approximately 2.5× compared to the clean environment. These evaluation results clearly show that the introduced sim-to-real perturbations substantially affect navigation performance. In contrast, the performance of \textit{NavRL++} (Table \ref{tab:benchmark_results}) demonstrates that perturbation-aware fine-tuning effectively mitigates degradations and improves robustness under realistic conditions.

In addition to analyzing sim-to-real discrepancies, we evaluate how practical deployment factors, specifically velocity limits and control frequency, affect performance. Using \textit{NavRL++}, we vary these parameters and report the results in Table \ref{tab:random_goal_vel_freq}. The results show that increasing or slightly reducing the velocity limit has negligible impact in static environments, while lowering it to 1.5 m/s leads to a modest performance drop in dynamic settings. Overall, the policy remains robust across a reasonable velocity range, indicating flexibility for real-world deployment. Similarly, performance is largely preserved when the control frequency remains above 10 Hz. A frequency of 5 Hz causes only a minor degradation, whereas reducing it to 2 Hz significantly lowers the navigation success rate. These results demonstrate that \textit{NavRL++} maintains strong performance under varied deployment conditions.

\begin{table}[t]
\centering
\caption{Comparison of cross-simulator evaluation performance in success rate (SR, \%) and trajectory effort (TE, $m/s^2$).}
\label{tab:gazebo_benchmark}

\renewcommand{\arraystretch}{1.2}
\setlength{\tabcolsep}{2.5pt}
\begin{tabular}{l cc cc cc}
\toprule
\multirow{2}{*}{\textbf{Method}}
& \multicolumn{2}{c}{\textbf{Low-Complexity}}
& \multicolumn{2}{c}{\textbf{Mid-Complexity}}
& \multicolumn{2}{c}{\textbf{High-Complexity}} \\
\cmidrule{2-3}\cmidrule{4-5}\cmidrule(lr){6-7}
& SR [\%] $\uparrow$ & TE $\downarrow$
& SR [\%] $\uparrow$ & TE $\downarrow$
& SR [\%] $\uparrow$ & TE $\downarrow$\\
\midrule

\multicolumn{7}{l}{\textbf{Static Env.}} \\
\addlinespace[1pt]

Fast-Planner \cite{fast-planner} & 99 & 0.110 & \textbf{100} & 0.120 & 96 & 0.120 \\
EGO-Planner \cite{ego-planner}  & 99 & \textbf{0.080} & \textbf{100} & \textbf{0.082} & 99 & \textbf{0.082} \\
ViGO \cite{ViGO}                & 91 & 0.084 & 89  & 0.093 & 86 & 0.087 \\
NavRL \cite{NavRL}              & 83 & 0.120 & 79  & 0.124 & 75 & 0.132 \\
\rowcolor{gray!20}
\textbf{NavRL++ (Ours)}          & \textbf{100} & 0.130 & 99 & 0.130 & \textbf{100} & 0.130 \\

\midrule

\multicolumn{7}{l}{\textbf{Dynamic Env.}} \\
\addlinespace[1pt]

ViGO \cite{ViGO}   & 71 & \textbf{0.088} & 70 & \textbf{0.091} & 57 & \textbf{0.103} \\
NavRL \cite{NavRL} & 68 & 0.130 & 65 & 0.137 & 50 & 0.139 \\
\rowcolor{gray!20}
\rowcolor{gray!20}
\textbf{NavRL++ (Ours)} & \textbf{94} & 0.116 & \textbf{85} & 0.122 & \textbf{81} & 0.112 \\

\bottomrule
\end{tabular}
\end{table}

\begin{table}[t]
\centering
\setlength{\tabcolsep}{5pt}
\renewcommand{\arraystretch}{1.2}
\caption{Cross-simulator evaluation of path length in static long-distance (48 $m$) navigation scenarios.}
\label{tab:random_goal_static_gazebo}

\begin{tabular}{lccccc}
\toprule
\textbf{Path Length [m] $\downarrow$} 
& \textbf{Low} 
& \textbf{Mid} 
& \textbf{High} 
& \textbf{Average} 
& \textbf{Percentage} \\
\midrule

Fast-Planner \cite{fast-planner}
& 51.12 
& 51.25 
& 52.48 
& 51.62 
& 107.54\% \\

EGO-Planner \cite{ego-planner}
& \textbf{50.51} 
& \textbf{50.73} 
& \textbf{51.22} 
& \textbf{50.82} 
& \textbf{105.88\%} \\

ViGO \cite{ViGO}
& 50.56 
& 50.90 
& 51.89 
& 51.12 
& 106.50\% \\

NavRL \cite{NavRL}
& 52.58 
& 53.15 
& 52.85 
& 52.86 
& 110.13\% \\

\rowcolor{gray!20}
\textbf{NavRL++ (Ours)}
& 52.05 
& 52.47 
& 53.05 
& 52.52 
& 109.42\% \\

\bottomrule
\end{tabular}
\end{table}

\subsection{Cross-Simulator Evaluation \& Performance Benchmarking} \label{sec:cross-sim-results}
\begin{figure*}[t]
    \centering
    \includegraphics[scale=0.93]{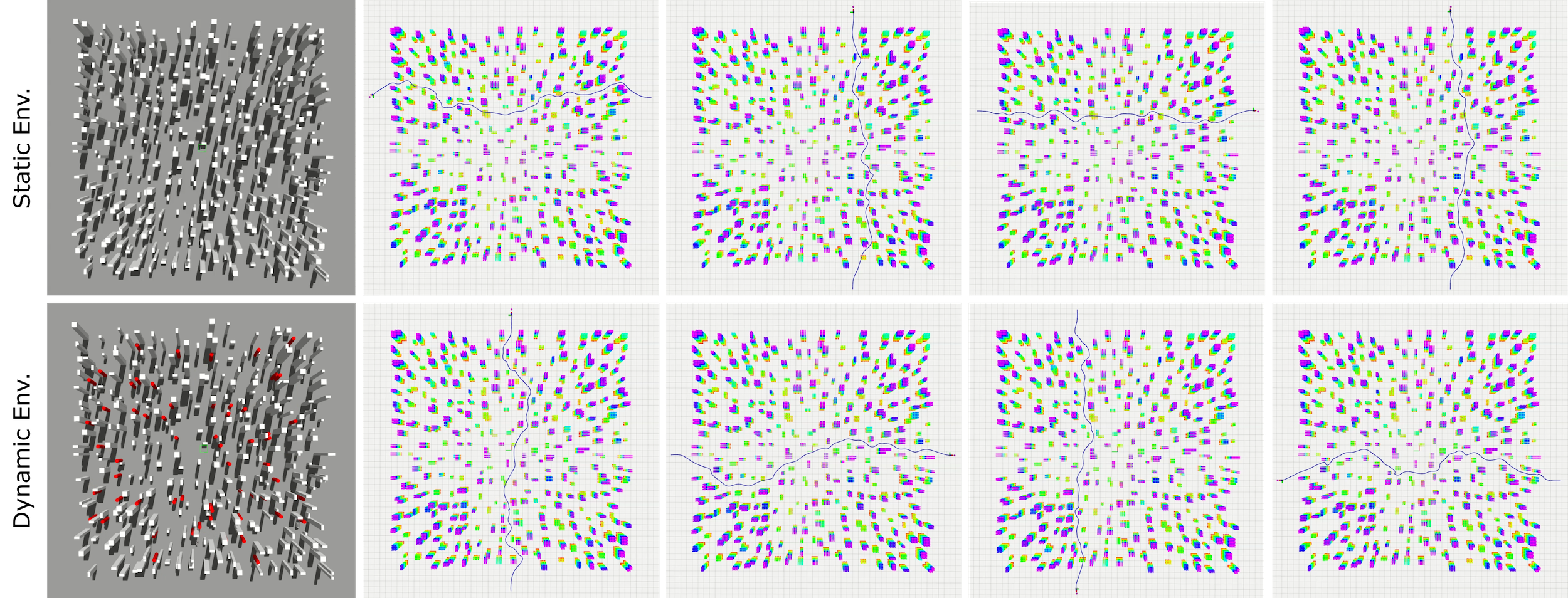}
    \caption{Gazebo navigation experiments with environment layouts and sample trajectories. The first column presents two randomly generated environments: one static and one dynamic, where static obstacles are in white and dynamic obstacles in red. The right four columns illustrate representative trajectories.}
    \label{fig:gazebo_sample_trajectory}
\end{figure*}

To evaluate generalization beyond the training simulator and benchmark against classical optimization based planners, we conduct cross-simulator experiments in Gazebo. Challenging static and dynamic obstacle environments are constructed (similar to how obstacles are placed in Isaac Sim), and each method is tested over 100 trials with randomly sampled start and goal positions. The example environments and results are shown in Fig. \ref{fig:gazebo_sample_trajectory}. We report the success rate and trajectory effort, where trajectory effort is defined as the time averaged norm of accumulated velocity changes, measured in $m/s^2$. Instead of control effort, trajectory effort is used because optimization based planners generate trajectories rather than direct control commands, and this metric reflects the smoothness and stability of the resulting motion. The results are summarized in Table \ref{tab:gazebo_benchmark}. We include representative planners designed for static environments, such as Fast-Planner \cite{fast-planner} and EGO-Planner \cite{ego-planner}, as well as methods that operate in both static and dynamic settings, including ViGO \cite{ViGO} and NavRL \cite{NavRL}. In static environments, our method achieves nearly 100 percent success across all scenarios, matching the performance of Fast-Planner and EGO-Planner. This consistency with the results observed in Isaac Sim further indicates strong cross simulator generalization. Regarding trajectory quality, EGO-Planner achieves the lowest trajectory effort, and optimization based methods generally produce smoother trajectories due to explicit predictive planning. This reveals a remaining limitation of learning based approaches in motion smoothness. However, in dynamic environments, our method achieves substantially higher success rates than the baselines, demonstrating superior robustness to moving obstacles. Overall, the proposed framework narrows the gap between reinforcement learning and optimization based navigation in static settings while delivering clear improvements in dynamic environments.

\begin{figure}[t] 
    \centering
    \includegraphics[scale=1.0]{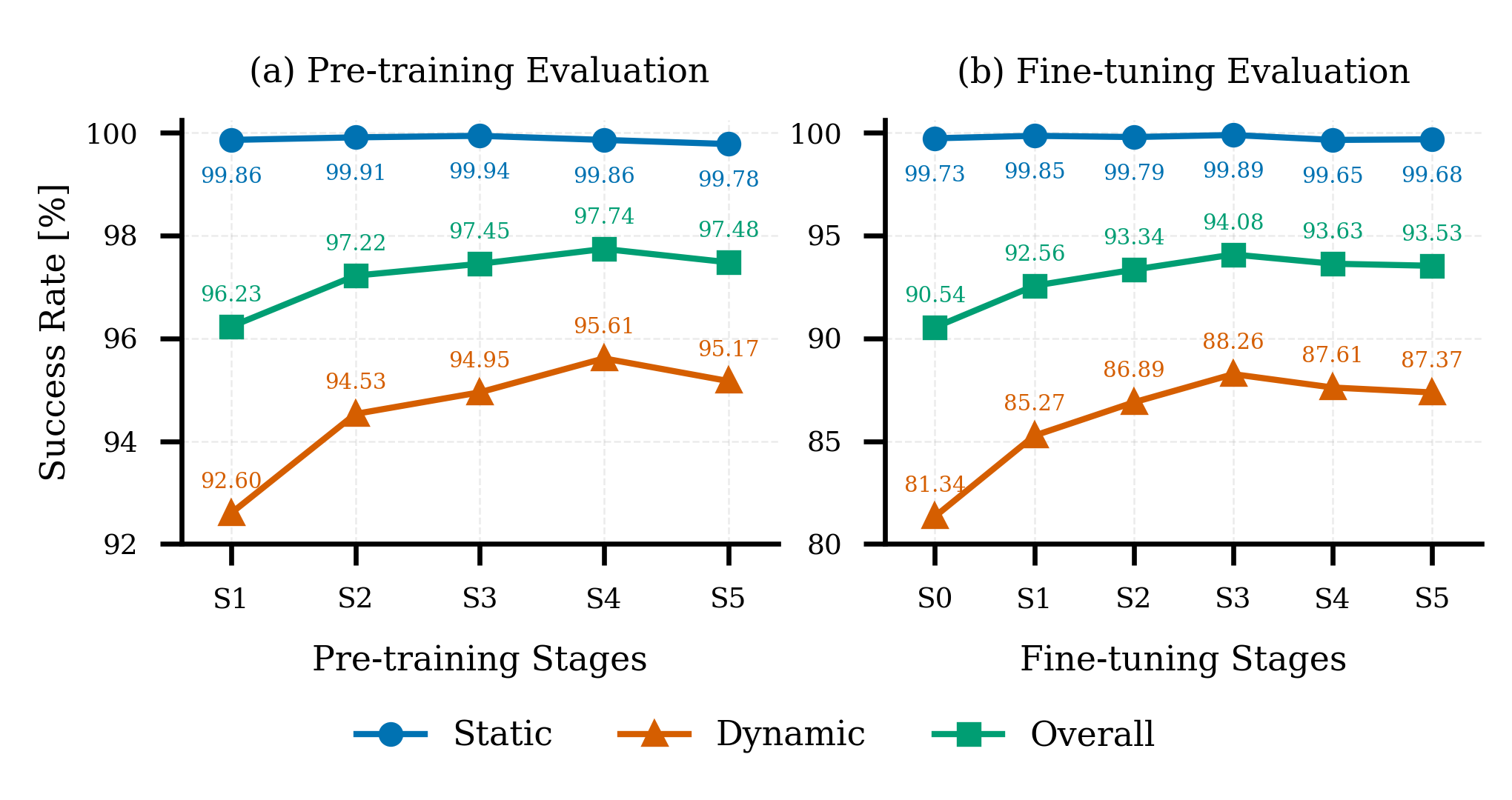}
    \caption{Navigation success rate evaluation at each curriculum training stage.}
    \label{fig:curriculum_learning}
\end{figure}

\begin{figure}[t] 
    \centering
    \includegraphics[scale=0.9]{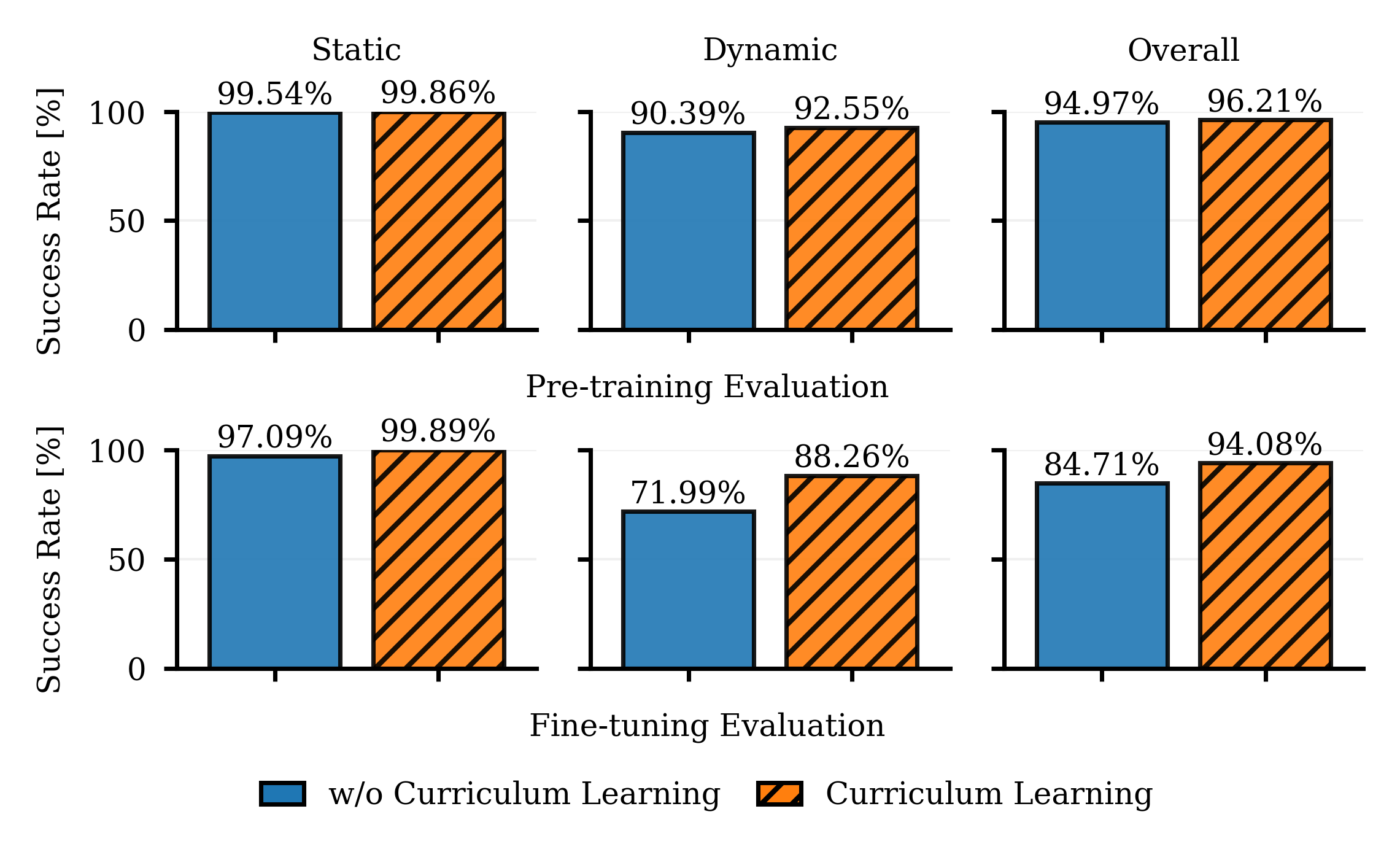}
    \caption{Comparison of navigation success rates between policies trained with and without curriculum learning in both static and dynamic environments.}
    \label{fig:curriculum_comparison}
\end{figure}

We further evaluate path length in long distance navigation scenarios with a fixed 48 meter separation between start and goal positions, consistent with the Isaac Sim setup. The results are presented in Table \ref{tab:random_goal_static_gazebo}. All methods achieve comparable path lengths within a 5\% variation range, which aligns with the trends in Isaac Sim. EGO-Planner produces the shortest paths, which is expected given its efficient trajectory optimization and low trajectory effort. Learning-based methods achieve slightly longer paths compared to optimization-based methods, but remain competitive overall. Together with the trajectory effort analysis, these findings suggest that improving trajectory efficiency and smoothness remains an important direction for advancing reinforcement learning-based navigation policies.

\subsection{Training Strategy \& Parameter Evaluation} \label{sec:training-parameters}
Reinforcement learning training strategies and parameters are important since inappropriate adjustments can prevent the policy from achieving satisfactory performance. Therefore, this section presents our empirical analysis of the most influential factors that significantly improve the performance and stability of the reinforcement learning-based navigation policy.

\textbf{Curriculum Learning.} Curriculum training is a widely used technique in reinforcement learning, and we evaluate its impact on navigation performance. We first record the navigation success rate at each curriculum stage, as shown in Fig. \ref{fig:curriculum_learning}, following the stages defined in Table \ref{tab:curriculum learning}, in both static and dynamic environments. The results indicate that the success rate generally improves as training progresses to later stages, with more noticeable gains observed in dynamic environments. However, performance does not increase monotonically with task difficulty. For example, the final pre-training stage (S5) and the last two fine-tuning stages (S4–S5) lead to slight performance degradation. To further verify the necessity of curriculum learning, we compare policies trained with and without curriculum learning, as shown in Fig. \ref{fig:curriculum_comparison}. The results demonstrate that curriculum learning improves the overall success rate by more than 1\% during pre-training and by approximately 10\% during fine-tuning. Notably, during fine-tuning evaluation, the success rate in dynamic environments improves by more than 15\%, which highlights the effectiveness of curriculum learning in complex settings.

\begin{figure}[t] 
    \centering
    \includegraphics[scale=0.85]{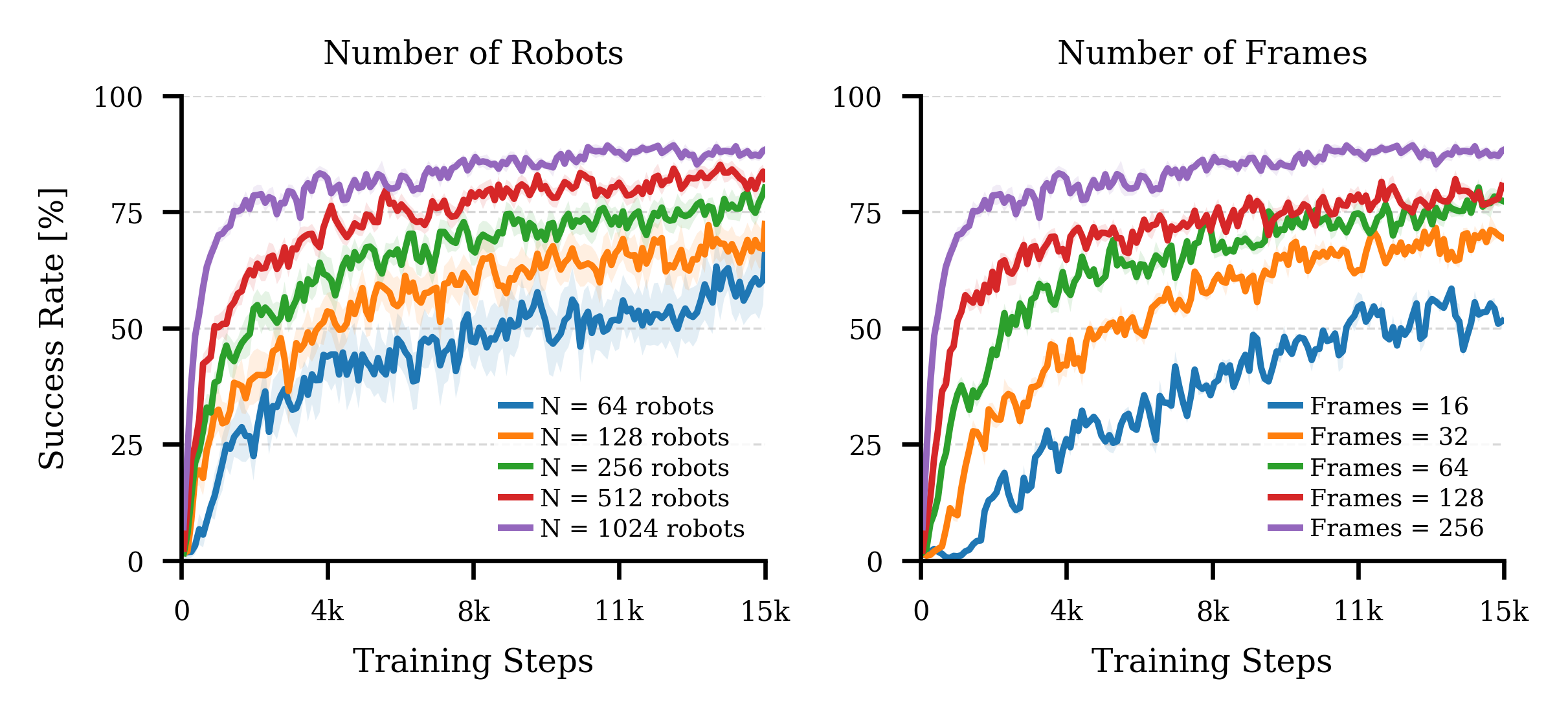}
    \caption{Comparison of training success rates under varying training scales. The results show that increasing both the number of robots running in parallel for data collection and the number of frames collected per policy update significantly improves training convergence speed and final success rate.}
    \label{fig:training_num_robot_and_frames}
\end{figure}

\begin{figure}[t] 
    \centering
    \includegraphics[scale=0.96]{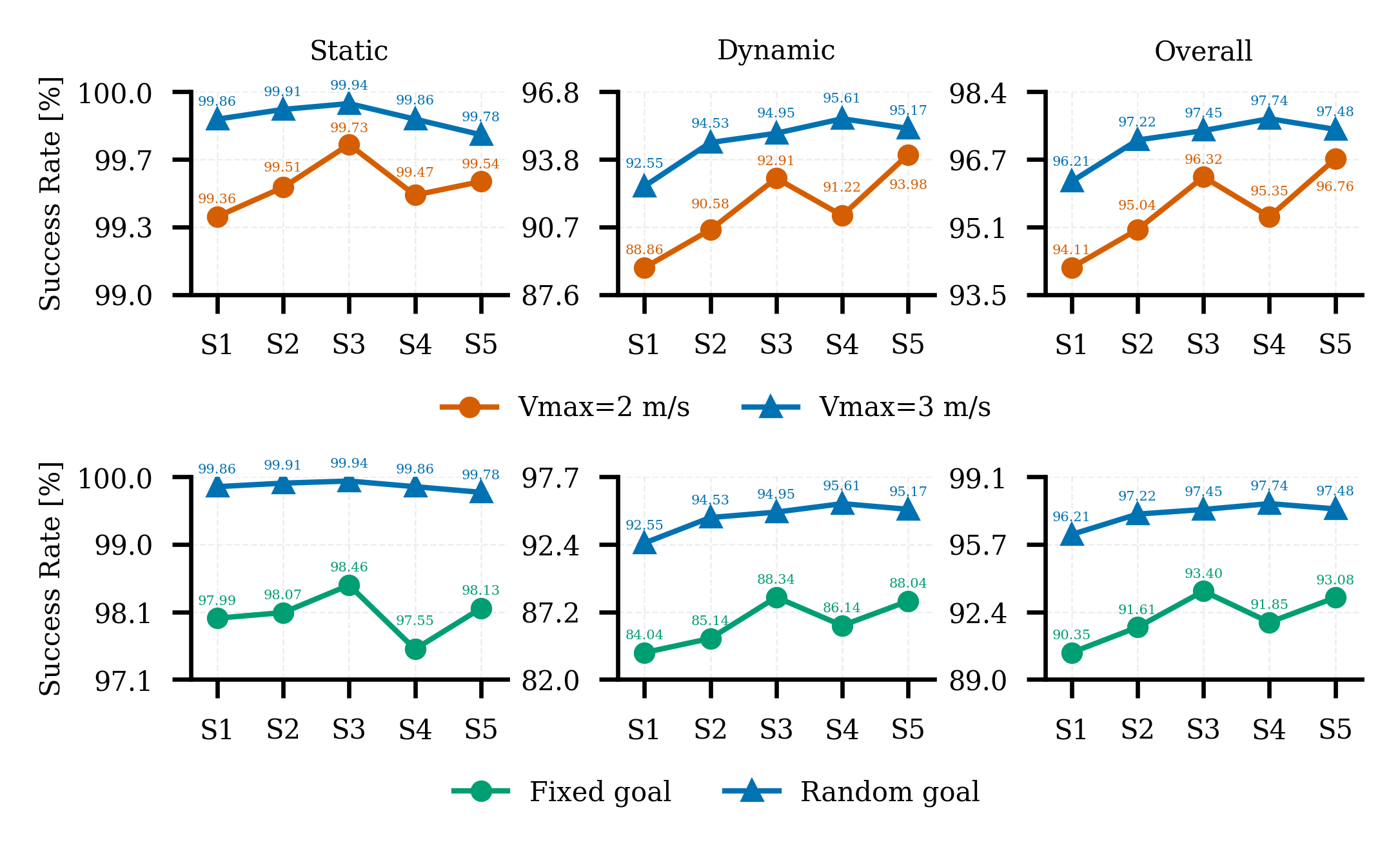}
    \caption{Comparison of evaluation success rates with and without goal randomization during each training stage. The results demonstrate that incorporating goal randomization significantly improves training success rates.}
\label{fig:vel_and_goal_experiments}
\end{figure}

\begin{figure}[t]
    \centering
    \includegraphics[scale=1.15]{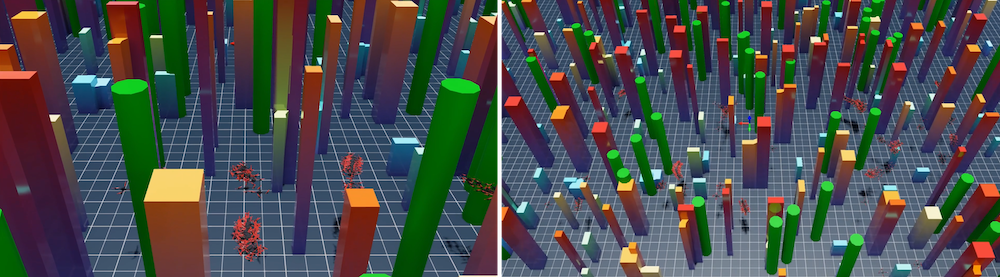}
    \caption{Navigation in Isaac Sim. Static obstacles are visualized as red and yellow objects and dynamic obstacles are represented by green cylinders.}
    \label{fig:isaac_sim_qualitative_experiments}
\end{figure}
\textbf{Large-Scale Training.} Another important factor is the training scale, which is determined by two key parameters: the number of robots deployed in parallel for data collection and the number of frames collected per update. Fig. \ref{fig:training_num_robot_and_frames} compares training success rates under different configurations and illustrates the learning curves for the first curriculum stage during pre-training. The left plot in Fig. \ref{fig:training_num_robot_and_frames} shows that increasing the number of robots accelerates convergence and improves final success rate, while the right plot in Fig. \ref{fig:training_num_robot_and_frames} demonstrates a similar trend when increasing the number of frames per update. These results highlight the importance of large-scale training in reinforcement learning-based navigation.

\textbf{Goal Randomization.} The prior method \cite{NavRL} trains the policy by sampling goal positions from a fixed set of predefined locations, which facilitates convergence but limits generalization. In our implementation, we randomly sample goal positions for each robot at every episode, regardless of whether the sampled goal is in collision. As shown in Fig. \ref{fig:vel_and_goal_experiments}, this modification significantly improves performance. The figure compares success rates across curriculum stages during both pre-training and fine-tuning in static and dynamic environments. The results clearly indicate that goal randomization consistently improves success rates under all settings.

\begin{figure}[t]
    \centering
    \includegraphics[scale=0.75]{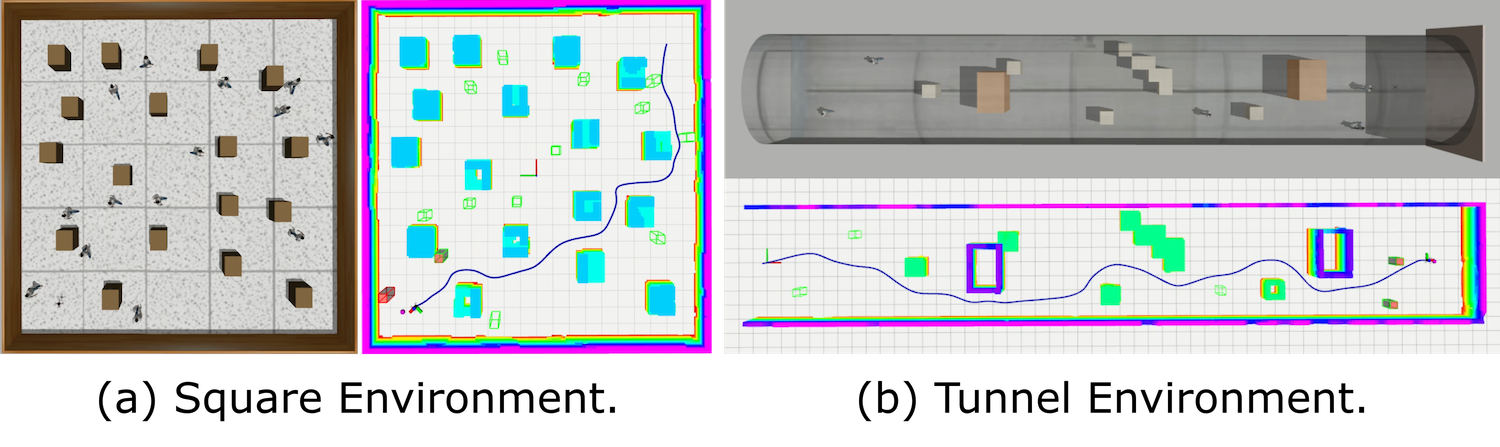}
    \caption{Navigation experiments in Gazebo simulated real-world scenarios.}
    \label{fig:gazebo_qualitative_experiments}
\end{figure}

\subsection{Qualitative Simulation Experiments} \label{sec:simulation-results}
Previous sections present quantitative results. In this section, we provide qualitative evaluations in simulation to offer an intuitive understanding of how the proposed method performs across different environments. Fig. \ref{fig:isaac_sim_qualitative_experiments} illustrates navigation in dynamic Isaac Sim environments, where multiple quadcopter UAVs operate in parallel without interfering with one another. In addition to Isaac Sim, we evaluate performance in Gazebo under similarly cluttered conditions. Fig. \ref{fig:gazebo_sample_trajectory} presents example static and dynamic environments, along with representative trajectories from multiple start–goal pairs, demonstrating successful collision-free navigation. Since these evaluations are designed to be highly challenging and mainly test the limits of the proposed method, we further assess performance in more realistic simulation settings, as shown in Fig. \ref{fig:gazebo_qualitative_experiments}. These include square and tunnel environments with static structures and moving pedestrians. In both scenarios, the robot successfully navigates the environments without collision. The full simulation evaluation can be found in the accompanying video.

\subsection{Real-World Evaluation Experiments} \label{sec:real-world-results}

\begin{figure*}[t]
    \centering
    \includegraphics[scale=0.94]{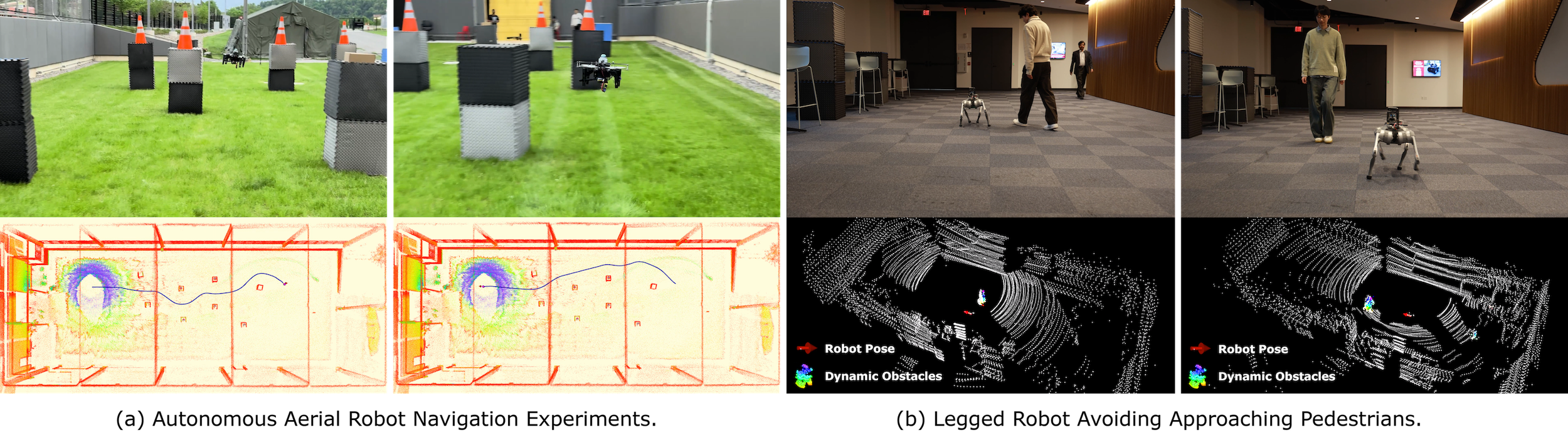}
    \caption{Real-world navigation experiments on both aerial and legged robotic platforms. (a) Navigation in cluttered environments using a UAV, with the trajectory and online map visualized. (b) Legged robot avoiding approaching pedestrians, with perceived dynamic obstacles highlighted in the RViz views.}
    \label{fig:navigation_real_experiments}
\end{figure*}

\begin{figure*}[t]
    \centering
    \includegraphics[scale=0.887]{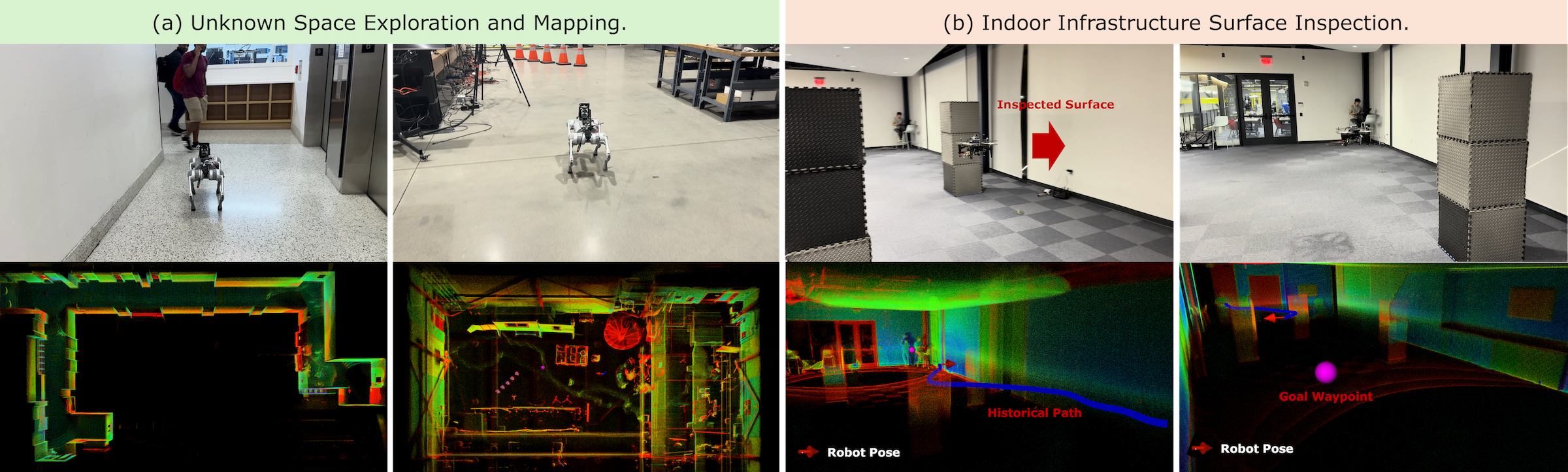}
    \caption{Navigation-centric applications using the proposed framework. In both exploration and inspection tasks, the task planner generates global waypoints that do not necessarily form a collision-free path from the robot’s position to the goal in the presence of unexpected obstacles. (a) Exploration and mapping of unknown environments, with the final explored maps shown. (b) Indoor infrastructure surface inspection, with the RViz view of the inspected surface.}
    \label{fig:application_experiments}
\end{figure*}

We further evaluate the proposed method in real-world environments. We begin with navigation experiments in both static and dynamic settings using multiple robotic platforms, including a quadcopter UAV and a quadruped robot. The same policy is applied across platforms without any platform-specific tuning. These experiments are shown in the top two columns of Fig. \ref{fig:intro_figure} and in Fig. \ref{fig:navigation_real_experiments}. In these scenarios, robots navigate through cluttered environments with pedestrians approaching from both the front and lateral directions. The results show that the proposed method enables safe navigation in crowded spaces while effectively avoiding dynamic obstacles, and that it generalizes well across different robot platforms.

To further demonstrate applicability to real-world tasks, we evaluate the proposed method in two navigation-centric applications, as shown in Fig. \ref{fig:application_experiments}. The first application is unknown environment exploration and mapping, illustrated in the left two columns of Fig. \ref{fig:application_experiments}. In this task, a robot operates without a prior map and must explore the environment while building a map using onboard sensors. A global exploration planner \cite{DEP} provides waypoint goals. These waypoints do not necessarily guarantee collision-free paths and may involve unseen obstacles, which requires local obstacle avoidance. The proposed method is used for local navigation to follow these global waypoints. Experiments conducted in multiple environments show that the robot maintains safe distances from obstacles and successfully builds the environment map.

The second application is indoor infrastructure surface inspection, shown in the right two columns of Fig. \ref{fig:application_experiments}. In this task, a sequence of waypoints generated from a floor plan guides the robot to traverse the environment and collect surface data. We deploy a quadcopter UAV and use the proposed method to navigate between these waypoints. Since the waypoints are generated from a prior floor plan, newly introduced obstacles are not considered during global planning. This again requires reliable collision avoidance. The results show that the robot successfully avoids obstacles and completes the task.

All computations are performed on the onboard computer. The policy runs at 20 Hz with an average inference time of 4.1 ms, demonstrating real-time performance. The complete real-world experiments are provided in the accompanying video.

\section{Limitation}
While the proposed framework demonstrates promising improvements and practical real-world deployment, we also discuss the limitations observed in our experiments.

\textbf{Perception.} The perception module converts multi-modal sensor inputs into a unified representation, which helps reduce the sim-to-real discrepancy and supports multiple sensing modalities. However, it relies on rule-based dynamic obstacle detection and tracking, which can lead to missed detections in complex environments and is sensitive to occlusion. These limitations may degrade navigation performance during deployment. Therefore, improving the perception module with more robust and adaptive methods remains an important direction for enhancing real-world navigation robustness.

\textbf{Action Smoothness.} From the evaluation comparison with state-of-the-art optimization-based methods, action smoothness remains the primary gap in our approach, indicating robot motion oscillations during deployment. Although the proposed temporal reasoning policy effectively improves smoothness, there is still room for further refinement to match the trajectory quality achieved by optimization-based planners.

\textbf{Collision in Dynamic Environments.} Although the proposed RL navigation achieves near 100\% success in static environments, collision rates remain non-negligible in high-density dynamic settings, which is critical for safety-sensitive applications. Therefore, improved training strategies and more effective designs for handling dynamic obstacles are necessary to further enhance robustness in complex environments.

\textbf{Safety Shield.} We adopt an action safety shield during real-world deployment. While it is effective in typical environments, it can become overly conservative in highly crowded scenarios, leading the robot to remain stationary when no feasible optimization solution is found. Incorporating more advanced control reachability analysis could alleviate this issue by balancing computational efficiency and conservativeness.

\textbf{Narrow Space.} The trained policy may struggle in narrow space, typically environments with less than 0.1–0.2 m clearance on each side, where the robot can exhibit oscillatory or indecisive behavior around its current position. We attribute this limitation to the relatively sparse raycasting-based representation of static obstacles. Increasing the density of raycasting beams during training could help alleviate this issue by providing more precise spatial information.

\section{Conclusion} 
This work investigated deployment robustness in reinforcement learning-based robot navigation through an integrated system spanning sensing, perception, and control. We systematically analyzed how deployment discrepancies arising from sensing uncertainty, perception failure, latency, and control response mismatch affect navigation behavior, and incorporated these perturbations into training through perturbation-aware adaptation. The results further suggest that incorporating short-horizon temporal reasoning improves robustness under delayed or imperfect observations while reducing high-frequency control oscillation during deployment. In addition, the empirical analysis highlights the importance of curriculum design, training scale, and perturbation-aware adaptation in reinforcement learning-based navigation. Experimental results indicate that perturbation-aware adaptation improves deployment robustness under simulated domain discrepancies. Cross-simulator evaluations further show that the proposed method achieves performance comparable to representative optimization-based planners in static environments while exhibiting improved robustness relative to reinforcement learning baselines in dynamic settings. Real-world deployment across navigation-centric tasks, including unknown environment exploration and autonomous inspection, further demonstrates the applicability of the learned policies under diverse robotic platforms and sensing configurations. Despite improved deployment robustness, trajectory smoothness and navigation in highly constrained environments remain challenging for reactive reinforcement learning policies. More broadly, this work highlights the importance of studying reinforcement learning deployment from a system-level autonomy perspective rather than through isolated architectural improvements alone.



\bibliographystyle{unsrtnat}

\bibliography{references}

\end{document}